\documentclass{article}
\usepackage{booktabs,array,tabularx}
\usepackage{caption}
\usepackage{multirow}
\usepackage{aaai2027_conference}
\usepackage[utf8]{inputenc}
\usepackage[T1]{fontenc}
\usepackage{hyperref}
\usepackage{xcolor}
\hypersetup{
    colorlinks=true,
    linkcolor=blue,
    urlcolor=blue,
    citecolor=blue,
    linkbordercolor=blue,
    urlbordercolor=blue,
    citebordercolor=blue,
    pdfborderstyle={/S/U/W 1},
}
\usepackage{float}
\usepackage{url}
\usepackage{booktabs}
\usepackage{amsfonts}
\usepackage{nicefrac}
\usepackage{microtype}
\usepackage{graphicx}
\usepackage{wrapfig}
\usepackage{natbib}
\usepackage{amsmath}
\usepackage{amssymb}
\usepackage{amsthm}
\usepackage{xspace}
\usepackage{enumitem}
\usepackage{multirow}
\usepackage{subcaption}
\usepackage{caption}
\DeclareCaptionLabelSeparator{pipe}{ | }
\usepackage{tcolorbox}
\newtcolorbox{coloredquote}[1][]{     
    colback=green!5!white,
    colframe=green!70!black,
    boxrule=2pt,
    arc=7pt,
    left=6pt,
    right=6pt,
    top=4pt,
    bottom=4pt,
    title=#1
}

\setlist[itemize]{leftmargin=*}
\setlist[enumerate]{leftmargin=*}
\setlist[description]{leftmargin=*}

\newcommand{\figorplaceholder}[2]{%
  \IfFileExists{#1}{\includegraphics[width=#2]{#1}}{%
    \fbox{\parbox[c][1.25in][c]{#2}{\centering Missing figure file:\\ \texttt{\detokenize{#1}}}}}%
}

\title{What RL Brings to Reasoning: A Few Decisions the Base Model Would Not Make}

\author{
Siyu Zhang\textsuperscript{1,2,3}\thanks{Equal contribution.} \thanks{Work done during an internship at Meituan LongCat Team.}, Yuquan Lu\textsuperscript{2,3}$^{*}$, Yoshua Y. Li\textsuperscript{2}$^{*}$\thanks{Corresponding author.} \\
\textbf{Yifan Ding\textsuperscript{2}, Kun Zeng\textsuperscript{3},  Yu Huo\textsuperscript{3}, Yifu Guo\textsuperscript{2} } \\
\textbf{Rongxiang Weng\textsuperscript{2},  Xunliang Cai\textsuperscript{2},  Xiaoying Tang\textsuperscript{3}$^{\ddagger}$} \\
\textsuperscript{1}University of California, San Diego \quad
\textsuperscript{2}Meituan, LongCat Team \\
\textsuperscript{3}The Chinese University of Hong Kong, Shenzhen \\
\texttt{siz036@ucsd.edu, yoshua\_li@meituan.com} \\
}

\renewcommand{\headeright}{}              
\renewcommand{\shorttitle}{What RL Brings to Reasoning}    

\begin{document}
\maketitle
\setcounter{footnote}{0}

\providecommand{\rlrep}{\textsc{rl-replace}}

\begin{abstract}
Reinforcement learning with verifiable rewards (RLVR) substantially improves mathematical reasoning, but what it adds to the base model remains unclear. We investigate this through a sequence of generation-time interventions on Qwen3-4B-Base, measuring how much of its RL-trained counterpart's gain each intervention recovers on 120 AIME problems. Power sampling with the same 8{,}192-token output limit brings no gain beyond greedy decoding. Yet borrowing the RL model's token only where the two models' greedy choices differ, roughly 4\% of generated positions, recovers 90.5\% of the pass@1 gain, with similar recovery across three model pairs on MATH-500. We find that even a single opening token has a substantial effect: forcing the base model to begin with ``Alright'' recovers 35\% of the RL gain. The opening alone is insufficient, however. Later replacements after the same opening add another 5.2 percentage points to pass@1, and the fixed rules we test for inserting reflective words such as ``Wait'' do not reproduce this benefit. Sparse replacement also retains much of the base model's diversity while staying close to the RL model at pass@1. At $k=256$, it solves 67.5\% of the problems, significantly more than the RL model (61.4\%) and a mixture of whole responses matched in pass@1 (61.45\%). The comparison is at equal attempt counts rather than equal compute, since rl-replace consults both models at every decoding step. These results show that the opening can recover a meaningful part of the RL gain, but a larger contribution comes from later choices, which can be borrowed while retaining much of the base model's diversity. Code is available at \href{https://github.com/CyrusZhang04/WhatRLBrings}{https://github.com/CyrusZhang04/WhatRLBrings}.
\end{abstract}

\section{Introduction}
\label{sec:intro}

Reinforcement learning with verifiable rewards (RLVR) has become a standard way to turn a pretrained language model into a strong mathematical reasoner~\citep{lambert2024tulu3,deepseek2025r1,shao2024deepseekmath}. Its performance gains are substantial, yet what RL adds to the base model remains unclear. Whether it teaches new reasoning abilities or mainly brings out ones the base model already has is still debated~\citep{yue2025doesrl,liu2025prorl,zhao2025echo}. Understanding this distinction requires identifying which changes in the model's behavior account for the improvement.

We start from differences visible in the generated responses. Our main comparison uses Qwen3-4B-Base and its GRPO-trained counterpart. On AIME, the RL model opens consistently with ``Alright, I have this problem\ldots'' and produces only 3.2 distinct openings among 16 responses, compared with 15.6 for the base model. It also reflects more and writes longer responses, using ``Wait'' about 5.7 times per response against 0.3 for the base model. These observations accompany broader findings that RL can sharply reduce entropy~\citep{cui2025entropy}, lose its advantage over the base model at large $k$~\citep{yue2025doesrl}, and concentrate its changes at a small set of high-entropy tokens~\citep{wang2025beyond,meng2026sparse}.

\begin{figure*}[t]
\centering
\includegraphics[width=\textwidth]{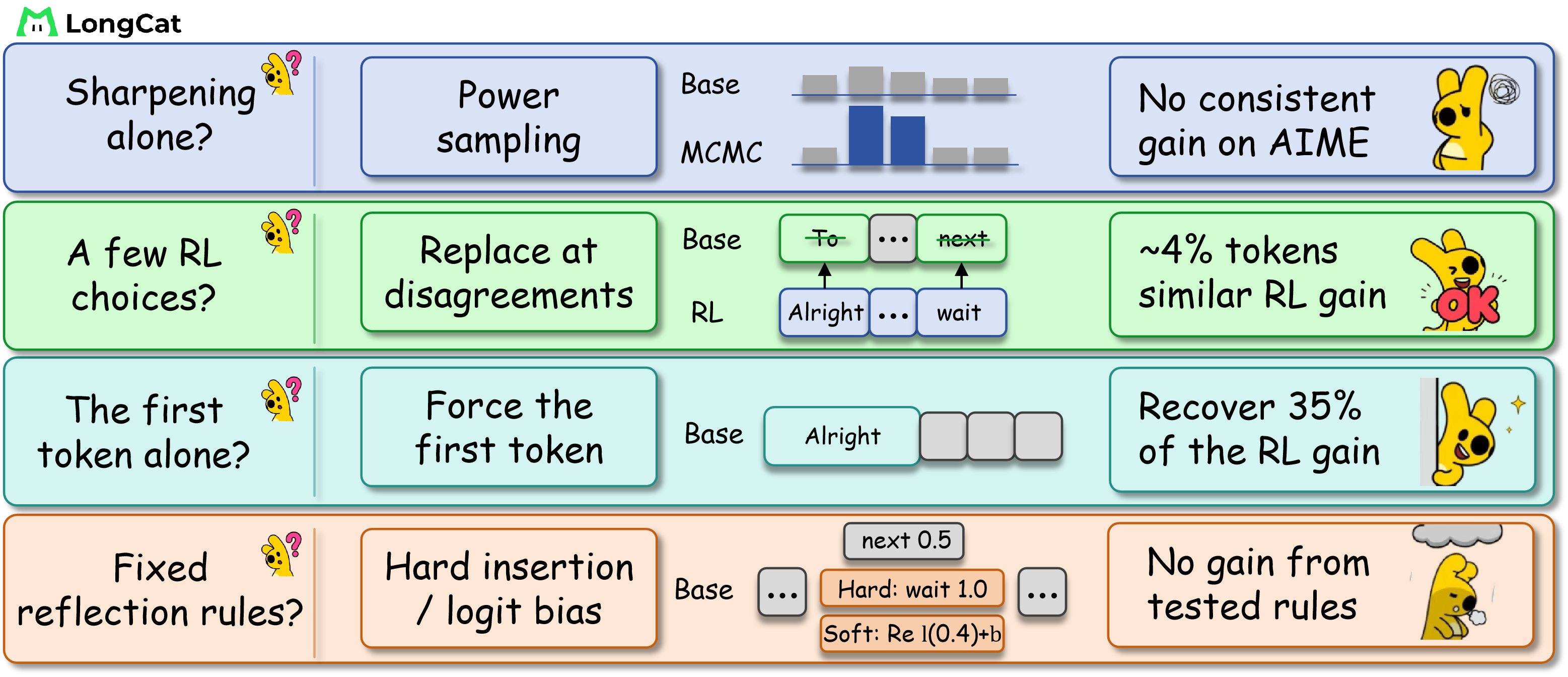}
\caption{Overview of the diagnostic tests of explanations for the RL gain. Each row pairs a question with its intervention and headline finding. All interventions act on the base model.
Tests~0 and~1 are evaluated on MATH-500 and AIME;
Tests~2 and~3 are evaluated on the 120 AIME problems.}
\label{fig:ladder}
\end{figure*}

These observations describe what changes, not which change produces the gain. Sharpening the base distribution, changing a few token choices, and altering the use of reflective behaviors have each been linked to improved reasoning~\citep{karan2026sampling,meng2026sparse,venhoff2026when}. However, evidence from different models, benchmarks and generation budgets does not establish how much each accounts for in the same setting. We therefore turn observed differences into interventions on the base model and ask five questions:
\begin{enumerate}
  \item Can power sampling reproduce the RL gain by sharpening the base distribution (Test~0)?
  \item Can most of the gain be recovered by changing only a small fraction of token choices (Test~1)?
  \item How much of this gain comes from the opening token, and how much from later choices (Test~2)?
  \item Can fixed rules over reflective tokens reproduce the benefit of the later choices (Test~3)?
  \item Can the RL accuracy gain be retained without the same loss of diversity?
\end{enumerate}

We study these questions on the 120 problems of AIME 2022--2025 using Qwen3-4B-Base and its GRPO-trained counterpart, where RL raises pass@1 from 10.00\% to 19.32\%. Under the tested configuration and the 8{,}192-token output limit, power sampling does not consistently outperform greedy decoding. A decoder we call \rlrep{} recovers 90.5\% of the RL gain by taking the RL model's greedy token where the two models' greedy choices differ, about 4\% of generated positions, and sampling from the base model elsewhere. The gap between power sampling and sparse replacement also holds across three model pairs on MATH-500.

Using \rlrep{} as a reference, we find that forcing the opening token ``Alright'' alone recovers 35\% of the RL gain, while later replacements after the same opening add another 5.2 percentage points to pass@1. Neither base-greedy replacement nor the tested reflective-token rules reproduce the full benefit. Sparse replacement also preserves greater output diversity than the RL model and achieves higher pass@256 than both the RL model and a response-level mixture matched in pass@1. A similar accuracy--coverage pattern holds for Qwen3-8B. These coverage comparisons use equal attempt counts; \rlrep{} consults both models at every decoding step.

Our contributions are as follows.
\begin{enumerate}
  \item We show that sparse token replacement can retain most of the RL accuracy gain alongside greater output diversity, while improving large-$k$ coverage over the RL model and a response-level mixture matched in pass@1.

  \item We quantify the contribution of the opening and the additional benefit of later RL choices, and show that base-greedy replacement and the tested reflective-token rules do not reproduce the full benefit.

  \item We provide a common intervention protocol for comparing explanations of the RL gain, and establish a substantial gap between power sampling and sparse replacement across three model pairs on MATH-500.
\end{enumerate}

\begin{figure*}[t]
    \centering
    \includegraphics[width=\textwidth]{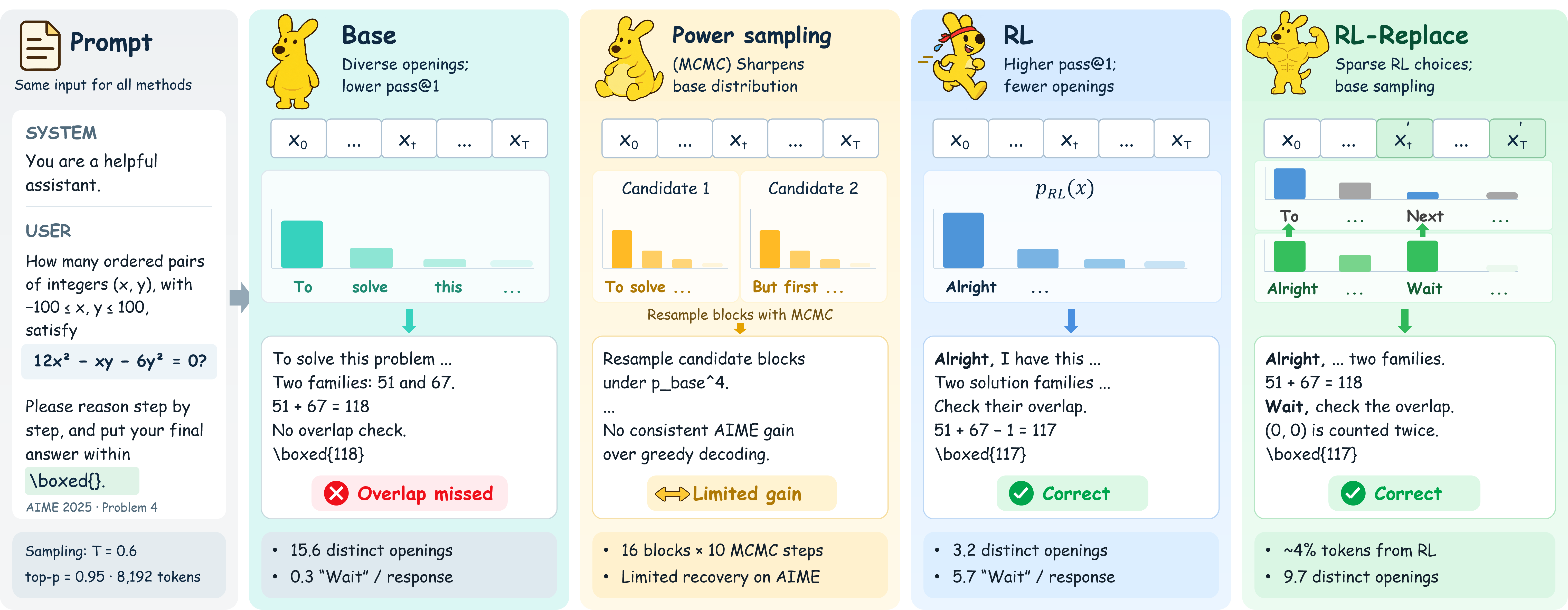}
    \caption{Illustrative comparison of the base model, power sampling,
    direct RL decoding, and sparse RL-token replacement on AIME 2025 Problem~4.
    All methods receive the same prompt. The base model misses the overlap
    between two solution families, while the RL and RL-replace examples check
    for it and obtain the correct answer. Power sampling sharpens the
    base-model distribution but does not consistently recover the AIME gain.
    The bottom annotations report the corresponding opening and token-level
    statistics.}
    \label{fig:method_comparison}
\end{figure*}

\section{Related Work}
\label{sec:related}

\paragraph{What does RL add to a base model?}
Reinforcement learning with verifiable rewards~\citep{lambert2024tulu3}, typically with GRPO~\citep{shao2024deepseekmath} or DAPO~\citep{yu2025dapo}, is now standard for improving mathematical reasoning~\citep{deepseek2025r1,kimi2025k15,zeng2025simplerl}, but whether it teaches new abilities remains debated~\citep{liu2025prorl,li2025questa,yuan2025composing}. Base models catch up with their RL-trained counterparts in pass@$k$ at large $k$~\citep{yue2025doesrl}, suggesting RL mainly makes the base model produce solutions it could already find~\citep{zhao2025echo,shao2025spurious,wang2025oneexample,matsutani2025squeezes,qin2025decomposing,wu2025leash}, at the cost of diversity~\citep{cui2025entropy,zhu2025negative,chen2025passk}. Reasoning behaviors such as backtracking and verification already exist in base models~\citep{gandhi2025cognitive,liu2025understanding}, and \citet{venhoff2026when} find thinking models mainly learn when to use them. Reflective tokens such as ``Wait'' carry answer information~\citep{qian2025infopeaks} and can extend thinking~\citep{muennighoff2025s1}, but suppressing them shortens outputs without lowering accuracy~\citep{wang2025wait}, and more reflection does not always help~\citep{kang2025firsttry,ghosal2025mirage,gema2025inverse,chen2024overthinking,wang2025underthinking,chen2025seal,yang2025aha}. We compare these explanations side by side, intervening on the base model of one RL model on hard competition problems.

\paragraph{Reasoning from the base model without training.}
A related line seeks RL-level reasoning from the base model directly~\citep{wei2022cot,wang2022selfconsistency,wang2024cotdecoding,agarwal2025entropy}. \citet{karan2026sampling} sample from the sharpened distribution $p_{\text{base}}^{\alpha}$ with MCMC and report results close to GRPO on MATH-500; later work changes where the sampler resamples~\citep{zhou2026cutting} and shows that sharpening can lower majority-vote accuracy~\citep{yang2026correctmass} and that Best-of-$N$ recovers most of its gains~\citep{talwar2026critical}. \citet{wang2026cue} show a short opening cue from training data, such as ``Alright'' for Qwen3, recovers much of the RL gain. We test both as explanations of the RL gain on hard problems at the RL model's generation budget (Tests~0 and~2).

\paragraph{Token-level analysis of RL.}
The effect of RL is sparse~\citep{mukherjee2025subnetworks} and concentrated on a few high-entropy decision tokens~\citep{wang2025beyond}, much as alignment changes only a few tokens~\citep{lin2023unlocking,zhou2023lima,qi2024safety}, and substituting the RL model's tokens at such positions recovers most of its accuracy~\citep{meng2026sparse,huang2026direction,shen2026restored,akgul2026sparse}. Our Test~1 uses such substitution, a decoding-time combination of two models~\citep{liu2024proxy,mitchell2023emulator,li2023contrastive,fei2024nudging,leviathan2022speculative}, as a reference rather than a finding. We ask how its gain divides between opening and later tokens, whether later tokens can be produced by fixed rules, and whether it keeps the base model's diversity.

\section{Analysis: Testing Explanations of the RL Gain}
\label{sec:analysis}

\begin{figure}[t]
\centering
\figorplaceholder{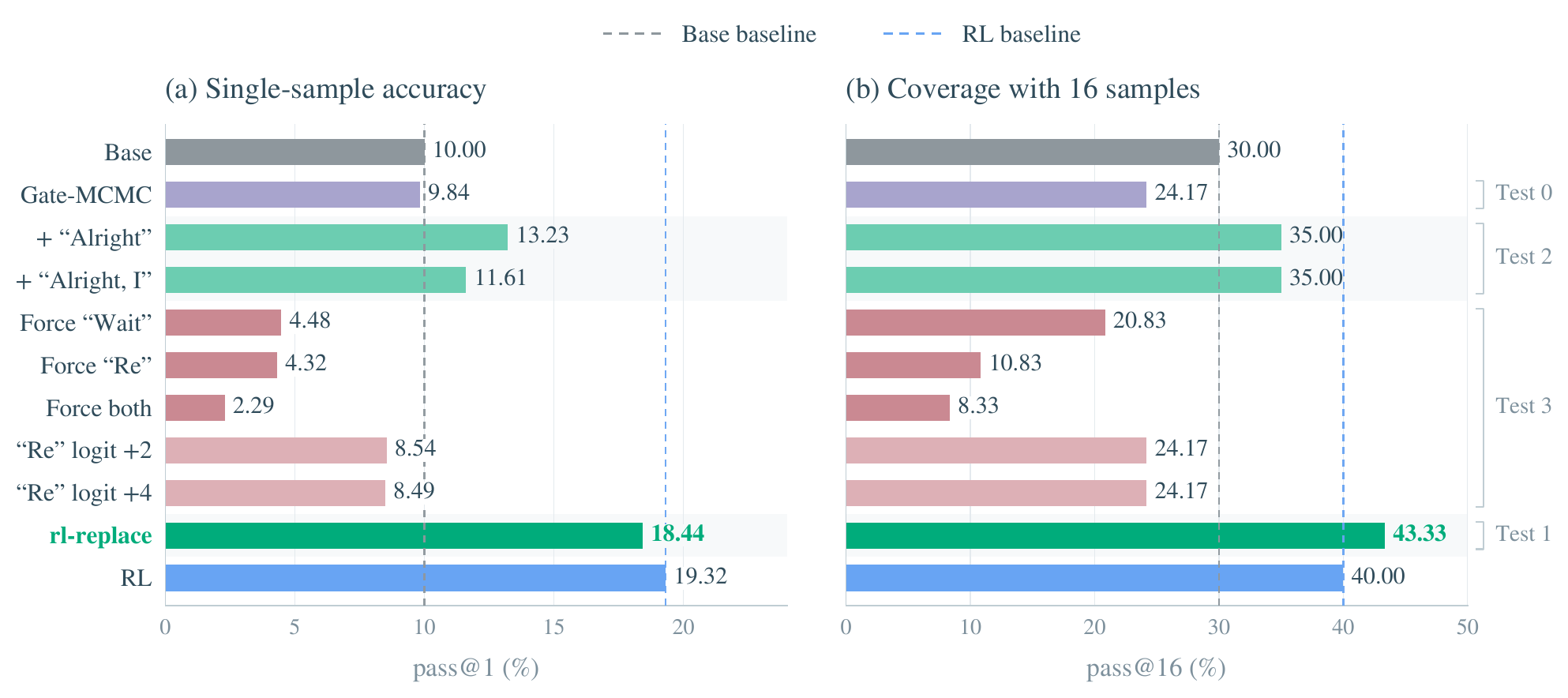}{\linewidth}
\caption{All interventions on the 120 AIME problems, with 16 samples per problem. Dashed lines mark the base model (grey) and the RL model (orange). Test~0 is represented by the gated MCMC variant of \S\ref{sec:sampling}.}
\label{fig:interventions}
\end{figure}

We compare interventions by the fraction of the RL pass@1 gain they recover, defined below. Figure~\ref{fig:ladder} summarizes the diagnostic tests, and Figure~\ref{fig:interventions} reports their results on AIME. Because intervention effects may overlap, recovery rates should not be interpreted as independent causal contributions that sum to the total gain. Section~\ref{sec:passk} then examines whether the recovered accuracy gain can coexist with greater output diversity and coverage at larger sampling budgets.

\subsection{Setup}
\label{sec:setup}

\paragraph{Models.}
Our analysis centers on Qwen3-4B-Base~\citep{yang2025qwen3} and an RL model trained from it directly, without supervised fine-tuning, using GRPO~\citep{shao2024deepseekmath} on DAPO-Math-17k~\citep{yu2025dapo} in verl~\citep{sheng2024hybridflow}. To avoid train–test contamination, we removed all problems duplicating AIME 2022 or 2023 from the DAPO training set prior to RL training. We keep the checkpoint with the highest accuracy on a validation split of DAPO-Math-17k (step 460), so no evaluation benchmark is used for checkpoint selection. RL raises MATH-500 pass@1 from 71.1\% to 84.4\% (Table~\ref{tab:mcmc}). Tests~0 and~1 are repeated on MATH-500 with Qwen3-8B-Base and Llama-3.2-3B~\citep{grattafiori2024llama3}, whose RL models are trained with the same recipe. Training hyperparameters are listed in Appendix~\ref{app:impl}.

\paragraph{Benchmarks and decoding.}
Our main benchmark is the 120 problems of AIME 2022--2025, reported for AIME24+25, AIME22+23 and all 120 problems; MATH-500~\citep{lightman2023verify} serves as an easier benchmark.  Every model and intervention receives the same prompt, which asks for the final answer in \texttt{\textbackslash boxed\{\}} (Appendix~\ref{app:prompts}). We sample at temperature 0.6 with top-$p$ 0.95 and at most 8{,}192 tokens, and grade the last boxed answer with Math-Verify~\citep{mathverify2025}.

\begin{figure}[t]
\centering
\figorplaceholder{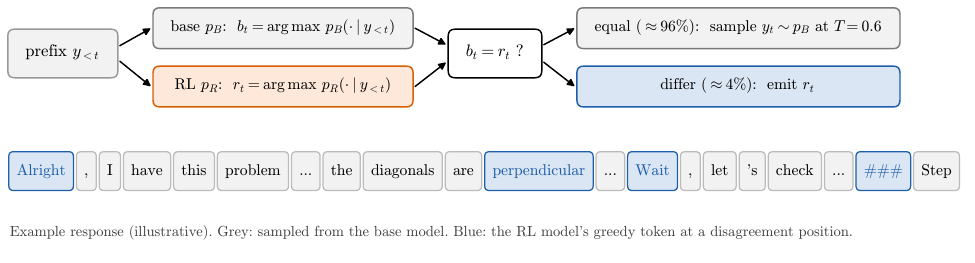}{\linewidth}
\caption{\rlrep{} decoding. At each step both models score the shared prefix; the RL model's greedy token is emitted only where the two greedy tokens differ (about 4\% of positions), and the base model samples everywhere else. The token sequence is illustrative.}
\label{fig:rlreplace}
\end{figure}

\paragraph{Sample budgets and metrics.}
On AIME we draw 16 samples per problem for every model and intervention, and on MATH-500 we draw 64. pass@1 is the mean accuracy over all samples, and pass@$k$ is the fraction of problems solved by at least one of $k$ samples, computed with the unbiased estimator of \citet{chen2021codex} when $k$ is smaller than the number of samples. The power-sampling comparison runs standard sampling and MCMC with 8 seeds and greedy decoding once, and the large-$k$ comparison of \S\ref{sec:passk} draws 512 samples per problem from the base and RL models and 256 from \rlrep. To compare interventions we use the \emph{recovery rate},
\begin{equation}
  R(\text{method}) = \frac{\text{pass@1}(\text{method}) - \text{pass@1}(\text{base})}{\text{pass@1}(\text{RL}) - \text{pass@1}(\text{base})},
\end{equation}
with 95\% confidence intervals from a problem-level bootstrap (10{,}000 resamples within each problem group). Differences between two conditions are resampled on the same problems (Appendix~\ref{app:stats}).

\paragraph{\rlrep.}
Test~1 relies on a decoder that we call \rlrep: whenever the two models prefer different next tokens, it takes the RL model's choice, and otherwise it keeps sampling from the base model. Formally, let $p_B$ and $p_R$ denote the base and RL models. At each step $t$ with prefix $y_{<t}$, we compute the two greedy tokens $b_t=\arg\max p_B(\cdot\mid y_{<t})$ and $r_t=\arg\max p_R(\cdot\mid y_{<t})$. If $b_t\neq r_t$ we emit $r_t$; otherwise we sample $y_t\sim p_B(\cdot\mid y_{<t})$ at temperature 0.6. The RL model therefore only intervenes at the positions where the two models' preferred next tokens differ, which we call \emph{disagreement positions}, and the base model writes everything else (Figure~\ref{fig:rlreplace}). Because both models run at every step, \rlrep{} costs 7.3--7.9 GPU-hours per AIME year, and we use it as a diagnostic of where the gain lies rather than as a practical decoding method. Appendix~\ref{app:impl} gives the full procedure and the settings of every experiment.

\subsection{Test 0: can power sampling recover the RL gain?}
\label{sec:sampling}

We first ask how much of the RL gain can be recovered by sharpening the base distribution alone. We use the power-sampling MCMC of \citet{karan2026sampling}, which targets $p_{\text{base}}^{\alpha}$ with $\alpha=4$, 16 blocks and 10 MCMC steps per block, following the original paper. All methods have the same 8{,}192-token output limit.

Table~\ref{tab:mcmc} shows two regimes. On MATH-500, MCMC improves pass@1 over standard sampling by 2.35--3.50 percentage points across all three base models, consistent with \citet{karan2026sampling}. For Qwen3-4B-Base, however, this improvement does not increase observed problem coverage with eight samples per problem. MCMC solves 400 of the 500 problems in at least one of its 8 seeds, compared with 408 for standard sampling (pass@8 of 80.0\% and 81.6\%). Only two problems are solved by MCMC alone, against ten by standard sampling alone. The higher pass@1 therefore mainly reflects more reliable success on problems also solved by standard sampling. Even on MATH-500, \rlrep{} and RL remain approximately 8--12 points above MCMC across all three models.

On AIME, MCMC provides no consistent gain over standard sampling and does not meaningfully outperform greedy decoding. For Qwen3-4B-Base, MCMC improves over standard sampling on AIME25 (13.75\% versus 9.58\%), but reaches essentially the same accuracy as greedy decoding (13.33\%). On AIME24, it stays near standard sampling and below greedy decoding. For Qwen3-8B-Base, MCMC falls below both standard sampling and greedy decoding on both benchmarks: it obtains 16.67\% on AIME24 and 20.42\% on AIME25, compared with greedy accuracies of 20.00\% and 23.33\%. Across the four model--benchmark pairs, \rlrep{} and RL exceed MCMC by approximately 5--14 points. Supplementary experiments on AIME 2022--2024 also show MCMC matching standard sampling at 10.00\% (Appendix~\ref{app:mcmc}).

The long generations expose additional difficulties for the sampler. Relative to standard sampling, the fraction of AIME responses with no final answer rises from 12.2\% to 22.4\%, and the fraction ending in repetition rises from 6.7\% to 13.1\%. Restricting MCMC to positions that a learned gate marks as likely high-KL positions (Appendix~\ref{app:predictor}) does not help either: on the 120 AIME problems, this variant obtains 9.84 / 24.17 pass@1 / pass@16, compared with 10.00 / 30.00 for standard base sampling.

These results identify a limit of power sampling under the tested configuration and output-length limit. Sharpening the base distribution improves pass@1 on MATH-500, but does not recover the RL gain on AIME for either Qwen model size. Applying the sampler selectively at predicted high-KL positions also brings no gain. This leaves a more specific question: how much of the gain can be recovered by borrowing the RL model's choices at only a small fraction of positions?

\subsection{Test 1: is the gain localized at a few positions?}
\label{sec:sparse}

We test this with \rlrep{}, which borrows the RL model's choice only where the two models' greedy tokens differ. On all 120 problems, it raises pass@1 from 10.00\% to 18.44\%, against 19.32\% for the RL model, recovering 90.5\% of the gain (Table~\ref{tab:main}). The recovery is similar on AIME24+25 (92.1\%) and AIME22+23 (87.7\%); the difference of 4.4 points has a 95\% interval of $[-32, +41]$ and is not significant.

The intervention is sparse. The RL model's token is used at about 4.0\% of all generated tokens (4.5\% when averaged per response), and the remaining 96\% are written by the base model. On MATH-500, \rlrep{} recovers the full RL gain (106.6\%), and it recovers most of the gain for the two other model pairs as well (\S\ref{sec:exp-models}). \citet{meng2026sparse} report a similar finding with a divergence-based cross-sampling procedure.

Sparse describes how many tokens are borrowed, not how far their effect reaches. Each borrowed token becomes part of the prefix the base model continues from, so an early choice can change everything the base model writes after it. This shows in the length of the responses: \rlrep{} responses average 6{,}217 tokens, close to the RL model's 6{,}504 and nearly twice the base model's 3{,}248 (Appendix~\ref{app:rep-all}). A few local decisions are thus enough to redirect the whole response, while every other token is still sampled from the base model. We use \rlrep{} as the reference for the next two tests. The first replacement is especially conspicuous: every response begins with the same word. We next isolate how much this opening contributes before examining the later choices.

\subsection{Test 2: how much does the opening contribute?}
\label{sec:firsttoken}

Every one of the 1{,}920 \rlrep{} responses on the 120 problems starts with ``Alright'', the first word of the RL model's usual opening, ``Alright, I have this problem\ldots''. The base model opens in varied ways, often by computing immediately or by continuing the chat template. This consistent first-token replacement suggests a simple intervention: force the base model to begin with ``Alright'' (or ``Alright, I'') and let it generate the rest without any help from the RL model.

Comparing this opening-only intervention with the base model measures what the cue itself contributes. Comparing it with \rlrep{}, which starts with the same word, measures the additional benefit of the later replacements. A third condition uses the same opening and disagreement rule as \rlrep{} but emits the base model's own greedy token at disagreements, allowing us to compare the benefit of committing to a token with that of using the RL model's choice.

\begin{table*}[t]
\centering
\caption{Power sampling (MCMC) versus standard sampling and greedy decoding, with methods that use the RL model as a reference. All values are pass@1 in \%. Greedy decoding uses a single run; sampling and MCMC are averaged over 8 seeds. Standard sampling uses temperature 0.6. 
For Qwen3-4B, the \rlrep{} and RL results on AIME come from
the main run summarized in Table~\ref{tab:main}, where standard
base sampling gives 12.50\% on AIME24 and 9.79\% on AIME25.}
\label{tab:mcmc}
\small
\begin{tabular}{@{}llccccc@{}}
\toprule
 & & \multicolumn{3}{c}{Base model only}
   & \multicolumn{2}{c}{Uses the RL model} \\
\cmidrule(lr){3-5}\cmidrule(l){6-7}
Benchmark & Model & Greedy & Sampling & MCMC & \rlrep{} & RL \\
\midrule
\multirow{3}{*}{MATH-500}
 & Qwen3-4B-Base & 71.80 & 71.10 & 73.45 & 85.25 & 84.37 \\
 & Qwen3-8B-Base & 74.60 & 75.98 & 78.53 & 87.74 & 89.26 \\
 & Llama-3.2-3B  & 46.80 & 46.15 & 49.65 & 57.50 & 59.27 \\
\midrule
\multirow{2}{*}{AIME24}
 & Qwen3-4B-Base & 16.67 & 10.00 & \phantom{0}9.58 & 23.13 & 23.13 \\
 & Qwen3-8B-Base & 20.00 & 20.00 & 16.67 & 27.08 & 26.67 \\
\midrule
\multirow{2}{*}{AIME25}
 & Qwen3-4B-Base & 13.33 & \phantom{0}9.58 & 13.75 & 21.04 & 22.92 \\
 & Qwen3-8B-Base & 23.33 & 20.83 & 20.42 & 25.42 & 26.25 \\
\bottomrule
\end{tabular}
\end{table*}

\begin{table*}[t]
\centering
\caption{\rlrep{} versus base and RL. AIME rows use $n=16$ samples per problem. Recovery is the share of the RL pass@1 gain recovered by \rlrep, with 95\% bootstrap intervals. On MATH-500, pass@1 of the base is standard sampling from Table~\ref{tab:mcmc} (temperature 0.6, averaged over 8 seeds), and pass@16 is the unbiased estimate from 64 samples per problem.}
\label{tab:main}
\small
\begin{tabular}{@{}lc ccc c ccc@{}}
\toprule
 & & \multicolumn{3}{c}{pass@1 (\%)} & & \multicolumn{3}{c}{pass@16 (\%)} \\
\cmidrule(lr){3-5}\cmidrule(lr){7-9}
Problems & \# & Base & \rlrep & RL & Recovery & Base & \rlrep & RL \\
\midrule
AIME24+25 & 60 & 11.15 & 22.08 & 23.02 & 92.1 [75.6, 121.2] & 36.67 & 48.33 & 41.67 \\
AIME22+23 & 60 & \phantom{0}8.85 & 14.79 & 15.63 & 87.7 [63.8, 120.9] & 23.33 & 38.33 & 38.33 \\
All & 120 & 10.00 & 18.44 & 19.32 & 90.5 [77.0, 108.6] & 30.00 & 43.33 & 40.00 \\
\midrule
MATH-500 & 500 & 71.10 & 85.25 & 84.37 & 106.6 [95.2, 120.4] & 82.86 & 94.46 & 94.65 \\
\bottomrule
\end{tabular}
\end{table*}

Table~\ref{tab:firsttoken} gives three findings. First, a single opening token recovers a substantial part of the RL gain. Forcing ``Alright'' raises pass@1 from 10.00\% to 13.23\%, recovering 35\% of the gain, and raises pass@16 from 30.00\% to 35.00\%. Extending the prefix to ``Alright, I'' adds no further improvement. In our setting, the opening provides a simple way to recover part of the gain, while leaving a substantial gap to the RL model.

Second, the later replacements contribute even more. Beyond the same opening, \rlrep{} adds 5.2 percentage points to pass@1 (95\% CI $[3.1, 7.6]$), equivalent to 56\% of the RL gain. The opening and later replacements together recover 90.5\%, compared with 35\% from the opening alone.

Third, choosing the base model's own greedy token recovers only a small part of this additional benefit. At a disagreement position, \rlrep{} both stops sampling and uses the RL model's token. Applying the same rule but emitting the base model's greedy token gives 14.74\%, 1.5 points above the opening alone and 3.7 below \rlrep{}. The larger gain therefore comes from using the RL model's choices, beyond the benefit of committing to a token under the same disagreement rule.

These comparisons measure the later replacements after an ``Alright'' opening; their benefit may include interactions with the opening, which we do not separate. Having established that later choices add more than the opening alone, we next examine what these choices look like and whether fixed rules can reproduce their benefit.

\subsection{Test 3: can fixed reflective-token rules recover the gain?}
\label{sec:rules}

\paragraph{What the later replacements look like.}
We examine the 19 problems on which the base model fails all 16 samples while \rlrep{} solves at least one. The 38 successful \rlrep{} responses contain 9{,}099 replacements that change the text (the rest only change tokenization). Table~\ref{tab:categories} groups them by type, and Appendix~\ref{app:replacements} gives the per-problem details. Most (74.7\%) are ordinary content-word changes, such as paraphrases and wording; opening and pause words such as ``Wait'', together with the prefix ``Re'', make up only 2.6\%. We focus the rule-based tests on these visible tokens because they are easy to imitate and often associated with reflective behavior after RL~\citep{deepseek2025r1,gandhi2025cognitive}.

Replacements occur throughout the response but are more frequent near the beginning: the first 20\% contains 26.1\% of all replacements, while the 60--70\% segment contains the smallest share (7.8\%). The category distribution is almost identical on the problems the RL model solves on its own and on those it does not (all categories within 0.5 points).

\begin{table*}[t]
\centering
\small
\begin{minipage}[t]{0.43\textwidth}
\centering
\caption{Forcing the RL model's opening on the base model (120 AIME problems, $n=16$, \%). The RL model is never consulted in the ``Alright'' and ``Alright, I'' rows; in the greedy row it only locates the disagreement positions, and the base model's own greedy token is used there.}
\label{tab:firsttoken}
\setlength{\tabcolsep}{4pt}
\begin{tabular}{@{}lcc@{}}
\toprule
Condition & pass@1 & pass@16 \\
\midrule
Base & 10.00 & 30.00 \\
Base + ``Alright'' & 13.23 & 35.00 \\
Base + ``Alright, I'' & 11.61 & 35.00 \\
Base + ``Alright'', base greedy & 14.74 & 36.67 \\
\midrule
\rlrep{} (reference) & 18.44 & 43.33 \\
RL (reference) & 19.32 & 40.00 \\
\bottomrule
\end{tabular}
\end{minipage}
\hfill
\begin{minipage}[t]{0.55\textwidth}
\centering
\caption{Replacements made by \rlrep{} in 38 successful responses on the 19 problems where the base model fails all 16 samples. Bidirectional categories count both directions.}
\label{tab:categories}
\setlength{\tabcolsep}{4pt}
\begin{tabular}{@{}lrr@{}}
\toprule
Category & Count & Share (\%) \\
\midrule
Other content words (paraphrase, wording) & 6{,}794 & 74.7 \\
\LaTeX{} formatting (both directions) & 1{,}074 & 11.8 \\
First-person pronouns (both directions) & 434 & 4.8 \\
Numbers & 202 & 2.2 \\
Opening and pause words ($\to$Alright, So$\to$Wait) & 190 & 2.1 \\
Markdown headers (Let$\to$\#\#\#) & 185 & 2.0 \\
Punctuation & 177 & 1.9 \\
``Re-'' prefix (Correct$\to$Re) & 43 & 0.5 \\
\midrule
Total & 9{,}099 & 100.0 \\
\bottomrule
\end{tabular}
\end{minipage}
\end{table*}

A few recurring patterns stand out. The RL model writes ``Wait'' where the base model's greedy token is ``So'', ``But'' or a section header (So$\to$Wait 20 times, But$\to$Wait 8, \#\#\#$\to$Wait 30), as after a total in one successful response (RL token in brackets):
\begin{quote}\small
Total = 404 + 405 = 809.\\
\textbf{[Wait]}, but we need to ensure that we are not double-counting.
\end{quote}
It also restructures the solution with a new header (Let$\to$\#\#\# and But$\to$\#\#\#, 24 times), typically right after such a pause, and turns a conclusion into a check (So$\to$But, 20 times). Number replacements show no fixed preference and depend on context, for example changing a value that violates a constraint stated earlier in the problem. Appendix~\ref{app:cases-example} shows a response in which a \#\#\#$\to$Wait replacement is followed by an overlap check and the correct answer. These examples motivate a narrower test: can simple rules that insert the same reflective tokens recover some of the gain?

\paragraph{Inserting the same tokens by rule.}
We build rules from the tokens the RL model uses and apply them to the base model without consulting the RL model during generation (Table~\ref{tab:rules}). Hard rules force a token at a trigger position. The triggers were derived from statistics of where the two models disagree, using a predictor of high-KL positions that relies only on the base model's own confidence (Appendix~\ref{app:predictor}). The ``Wait'' rule fires when the base model emits a section header right after a formula block, and the ``Re'' rule uses an analogous trigger for the reflective prefix. Soft rules add a constant to the logit of ``Re'' instead of forcing it.

\noindent
\begin{minipage}[t]{0.48\columnwidth}
\vspace{0pt}
Every hard rule falls well below the base model, by 5.5--7.7 points on pass@1 and 9--22 points on pass@16, and combining the rules is worst. A fixed trigger applies the same edit at the same kind of position across samples, whether or not a check is needed. For example, a section header after a formula block can mark an ordinary transition; forcing a pause there can interrupt reasoning that was on track. Soft rules bring no improvement either. Their pass@1 is 1.5 points below the base model, a difference within the variation we observe between independent base-model runs (\S\ref{sec:exp-years}), and the two bias strengths give nearly identical results.
\end{minipage}
\hfill
\begin{minipage}[t]{0.48\columnwidth}
\vspace{-10pt}
\centering
\captionof{table}{Rule-based insertion of the RL model's reflective tokens into the base model (120 AIME problems, $n=16$, \%).}
\label{tab:rules}
\footnotesize
\setlength{\tabcolsep}{4pt}
\begin{tabular}{@{}llcc@{}}
\toprule
Type & Condition & pass@1 & pass@16 \\
\midrule
-- & Base & 10.00 & 30.00 \\
\midrule
\multirow{3}{*}{Hard} & Force ``Wait'' & 4.48 & 20.83 \\
 & Force ``Re'' & 4.32 & 10.83 \\
 & Force ``Wait'' and ``Re'' & 2.29 & \phantom{0}8.33 \\
\midrule
\multirow{2}{*}{Soft} & ``Re'' logit $+2$ & 8.54 & 24.17 \\
 & ``Re'' logit $+4$ & 8.49 & 24.17 \\
\midrule
-- & \rlrep{} (reference) & 18.44 & 43.33 \\
\bottomrule
\end{tabular}

\end{minipage}

Other rule-like interventions lead to the same conclusion (Appendix~\ref{app:ablations}). Forcing ``Wait'' whenever it appears among the base model's top-5 candidates lowers accuracy to 6.25\%, and all 282 responses that receive six or more insertions are wrong, even though the RL model itself uses about 5.7 ``Wait''s per response. Forcing ``Wait'' after an ``Alright'' opening collapses accuracy to 0.26\%. Raising the probability of first-person pronouns, decoding the first 100 tokens greedily or at low temperature, and stopping early based on entropy all leave accuracy at the base level. On a small set of the hardest problems, replacing the RL token at the disagreement positions with a random base candidate yields 0 correct responses out of 480, compared with 3 out of 480 when the RL token is used; we treat this last comparison as qualitative given its size.

Together, these results show that merely increasing the frequency of the RL model's reflective tokens via fixed rules does not reproduce the accuracy benefit of sparse replacement, even though the rules raise the frequency of 'Wait' well above the base model's own usage (0.3 per response). The gap suggests that what rl-replace transfers is not reflective tokens as such, but the right reflective tokens at the right positions—along with many non-reflective content-word changes. One possibility is that RL sharpens the model's sense of when its reasoning has gone wrong, rather than teaching it to reflect unconditionally. To test the more specific view that RL learns when to deploy existing behaviors~\citep{venhoff2026when}, one could retain only the reflective replacements at positions chosen by the RL model and compare them with a rule that fires equally often. Our negative result is limited to the fixed rules tested here; we do not evaluate learned or entropy-based selection schemes such as that of \citet{akgul2026sparse}.

\subsection{Is the gain tied to the loss of diversity?}
\label{sec:passk}

\begin{figure*}[!t]
\centering
\includegraphics[width=\textwidth]
{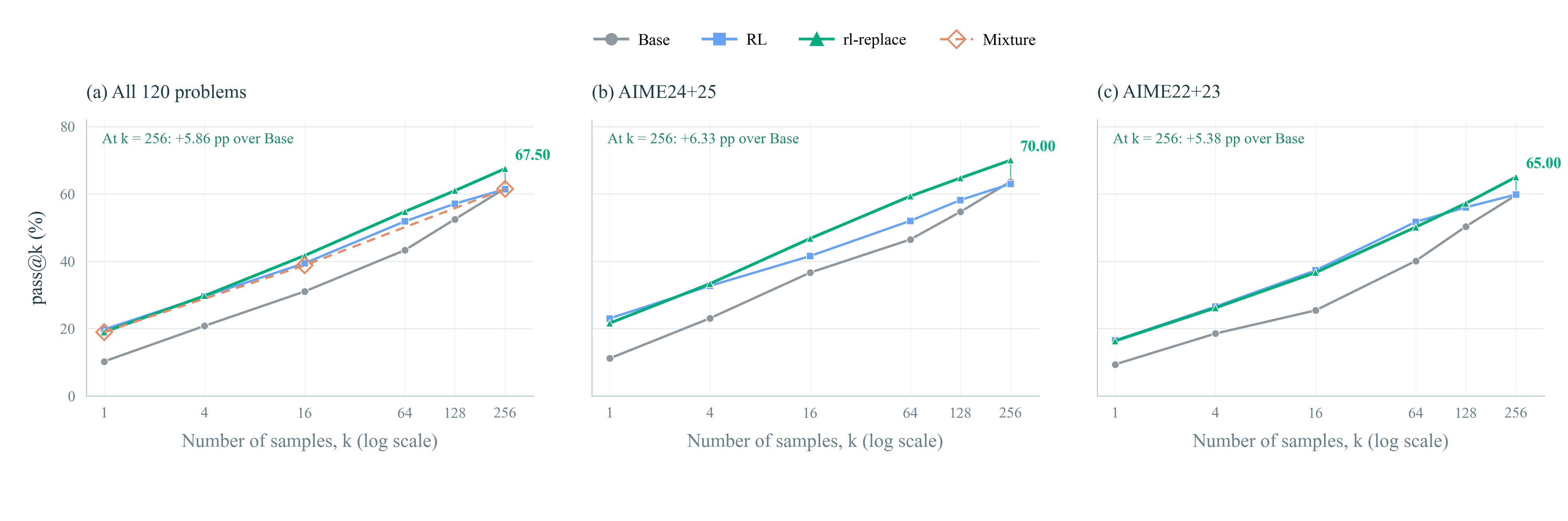}
\caption{Accuracy and coverage of Qwen3-4B as the sampling
budget increases: (a) all 120 AIME problems,
(b) AIME24+25, and (c) AIME22+23.
Estimates use 512 samples per problem for the base and
RL models and 256 for \rlrep{}, with the unbiased
pass@$k$ estimator.
The response-level mixture selects the RL model with
probability 0.92 and the base model otherwise for each
complete response; its dashed line connects the reported
values at $k=1,16,256$ in panel~(a).
All panels share the same vertical scale and use a
logarithmic horizontal axis.
Green endpoint labels report \rlrep{} pass@256.
The comparisons use equal attempt counts; \rlrep{}
consults both models at every decoding step.}
\label{fig:passk}
\end{figure*}

The preceding tests show that sparse replacement recovers
most of the RL accuracy gain. We now ask whether this
recovery also requires the RL model's loss of output
diversity. Table~\ref{tab:passk-main} reports the large-$k$
results for Qwen3-4B and Qwen3-8B, while
Figure~\ref{fig:passk} shows the Qwen3-4B curves for all
120 AIME problems and for the two problem groups.

\begin{table*}[!t]
\centering
\caption{pass@$k$ (\%) on all 120 AIME problems, computed
using the unbiased estimator with 512 samples per problem
for the base and RL models and 256 for \rlrep{}.
Bold indicates the highest value in each column within
each model group. The Qwen3-4B run is independent of the
main 16-sample run in Table~\ref{tab:main}, so its pass@1
and pass@16 values differ slightly.
Figure~\ref{fig:passk} shows the Qwen3-4B results by
problem group.}
\label{tab:passk-main}
\small
\setlength{\tabcolsep}{8pt}
\begin{tabular}{@{}lrrrrrr@{}}
\toprule
Method
& pass@1
& pass@4
& pass@16
& pass@64
& pass@128
& pass@256 \\
\midrule
\multicolumn{7}{@{}l}{\textit{Qwen3-4B}} \\
Base
& 10.26
& 20.79
& 31.02
& 43.29
& 52.50
& 61.64 \\
RL
& \textbf{19.77}
& 29.61
& 39.41
& 51.85
& 57.10
& 61.40 \\
\rlrep{}
& 18.98
& \textbf{29.74}
& \textbf{41.69}
& \textbf{54.73}
& \textbf{60.96}
& \textbf{67.50} \\
\midrule
\multicolumn{7}{@{}l}{\textit{Qwen3-8B}} \\
Base
& 19.77
& 27.31
& 38.64
& 50.18
& 58.09
& 68.33 \\
RL
& \textbf{28.32}
& \textbf{42.55}
& 50.10
& 57.35
& 60.64
& 66.82 \\
\rlrep{}
& 27.16
& 40.27
& \textbf{51.69}
& \textbf{58.31}
& \textbf{64.20}
& \textbf{72.50} \\
\bottomrule
\end{tabular}
\end{table*}

\paragraph{Accuracy and output diversity.}
For Qwen3-4B, \rlrep{} stays close to the RL model at
pass@1 in the large-$k$ run (18.98\% versus 19.77\%).
The diversity statistics from the main run, measured
over 16 responses per problem, show a different pattern.
The base model produces an average of 15.6 distinct
openings, defined as the first 300 characters, compared
with 3.2 for the RL model and 9.7 for \rlrep{}.
The corresponding numbers of distinct final answers are
9.42, 7.98, and 8.82.
Sparse replacement therefore retains greater output
diversity than the RL model while recovering most of
its accuracy gain.

\paragraph{Coverage at larger sampling budgets.}
Prior work finds that the RL model's advantage at pass@1
can disappear as $k$ grows~\citep{yue2025doesrl}.
Our Qwen3-4B pair follows this pattern:
the RL model leads the base model by 8.4--9.5 percentage
points from $k=1$ to $k=64$, but the lead shrinks to
4.6 points at $k=128$ and disappears at $k=256$
(61.40\% versus 61.64\%).

In contrast, \rlrep{} reaches 67.50\% at $k=256$
(Figure~\ref{fig:passk}a).
Its advantage over the RL model grows from 0.13 points
at $k=4$ to 2.28 at $k=16$ and 6.10 at $k=256$;
the last difference has a 95\% confidence interval
of $[0.82, 11.02]$.
Its observed advantage over the base model is
5.86 points, although the corresponding interval,
$[-0.33, 10.71]$, includes zero.
Thus, the large-$k$ results support a coverage advantage
over the RL model, while the comparison with the base
model remains less certain.

\paragraph{Comparison with a response-level mixture.}
Could mixing whole responses obtain the same combination
of accuracy and coverage?
We test a mixture that independently selects the RL model
with probability 0.92 and the base model otherwise for
each attempt.
The weight is chosen to approximately match \rlrep{}
at pass@1: 19.00\% versus 18.98\%.
The generation procedure is described in
Appendix~\ref{app:passk}.

The mixture remains close to the RL model, reaching
38.78\% at $k=16$ and 61.45\% at $k=256$.
At the latter budget, \rlrep{} leads by 6.05 percentage
points (95\% CI $[0.93, 10.98]$).
At approximately matched pass@1, combining the models
within each response therefore achieves higher pass@256
than the response-level mixture tested here.
These comparisons use equal numbers of attempts:
\rlrep{} runs both models at every decoding step,
whereas the mixture uses only one model per response.

\paragraph{Problem groups and model size.}
The Qwen3-4B coverage advantage over the RL model is
observed in both problem groups, although it emerges
at different budgets.
On AIME24+25, \rlrep{} leads by 5.2--7.3 points from
$k=16$ onward and reaches 70.00\% at $k=256$, compared
with 62.98\% for the RL model and 63.67\% for the base
model (Figure~\ref{fig:passk}b).
On AIME22+23, the RL model is slightly ahead through
$k=64$, but \rlrep{} overtakes it from $k=128$ onward,
reaching 65.00\% versus 59.81\% at $k=256$
(Figure~\ref{fig:passk}c).

Qwen3-8B shows a similar accuracy--coverage pattern
(Table~\ref{tab:passk-main}).
\rlrep{} remains close to the RL model at pass@1
(27.16\% versus 28.32\%) and achieves higher observed
pass@256: 72.50\%, compared with 66.82\% for the RL
model and 68.33\% for the base model.

\paragraph{Temperature sensitivity.}
Changing the Qwen3-4B RL model's temperature does not
close the observed coverage gap.
Among temperatures 0.4, 0.6, and 0.8 at fixed
top-$p=0.95$, the original setting of 0.6 gives the
highest RL pass@256.
The full results are reported in
Appendix~\ref{app:temperature}.

\paragraph{Implications for interpreting the RL gain.}
\citet{yue2025doesrl} interpret their coverage and reasoning-path analyses as evidence that current RLVR
primarily improves sampling efficiency over reasoning paths already available to the base model.
Our results highlight a distinction between the coverage of the RL model under its own decoding distribution and
the usefulness of its token choices within a different generation policy.
By selectively incorporating these choices, \rlrep{} recovers most of the RL accuracy gain while retaining greater output diversity and achieving higher pass@256 than the RL model and the response-level mixture tested
here.
This suggests that beneficial changes in token preferences can be retained without reproducing the full reduction in output diversity observed in the RL model.

These findings do not establish that RL creates reasoning paths unavailable to the base model: \rlrep{} is a hybrid
generation policy, and finite-budget pass@$k$ does not identify the full set of paths a model can generate.
Nor do they establish diversity loss as the cause of the coverage difference. Rather, they show that the RL model's own large-$k$ coverage does not fully characterize the value of the decisions it contributes to reasoning.

\section{Robustness}
\label{sec:robustness}

Our detailed intervention analysis centers on Qwen3-4B and a main run with 16 samples per problem on 120 AIME problems. We assess robustness across additional model pairs (Section~4.1), individual competition years and independent runs (Section~4.2). Larger sample budgets for both Qwen3-4B and Qwen3-8B are reported in Section~3.6.

\subsection{Other model pairs}
\label{sec:exp-models}

We repeat Tests~0 and~1 on MATH-500 with the two additional base/RL pairs, Qwen3-8B and Llama-3.2-3B, whose RL models are trained with the same recipe as the 4B model (\S\ref{sec:setup}); their pass@1 values are in Table~\ref{tab:mcmc}. The two tests give the same answer for all three pairs. Measured against standard base sampling, sharpening recovers 17.7\%, 19.2\% and 26.7\% of the RL gain for Qwen3-4B, Qwen3-8B and Llama-3.2-3B, while borrowing the RL model's choice at the disagreement positions recovers 106.6\%, 88.6\% and 86.5\%. The gap between Tests~0 and~1 is therefore not specific to Qwen3-4B, nor to a particular model family.

We also evaluate Qwen3-8B on AIME24 and AIME25 in the power-sampling comparison (Table~1), and on all 120 AIME problems in the large-$k$ comparison (Table~6). In the latter, \textsc{rl-replace} stays close to the RL model at pass@1 (27.16\% versus 28.32\%) while achieving higher observed pass@256 than either the base or RL model. These results extend the accuracy--coverage pattern to Qwen3-8B. The opening and reflective-token ablations in Sections~3.4--3.5 remain limited to Qwen3-4B, while Llama-3.2-3B is evaluated only on MATH-500.

\subsection{Individual years and independent runs}
\label{sec:exp-years}

The groups in Table~\ref{tab:main} pool two competition years each. Table~\ref{tab:years} breaks them down. \rlrep{} recovers between 85\% and 100\% of the RL pass@1 gain on every year, and its pass@16 is never below that of the RL model. The fraction of replaced tokens is stable across years (4.3--4.7\%), so the sparsity of the intervention does not depend on the problem set. Majority voting over 16 samples (maj@16) gives a noisier picture, as expected with 30 problems per year, but \rlrep{} improves over the base model on every year.

\begin{table*}[t]
\centering
\caption{Results by competition year ($n=16$, \%). Base / \rlrep{} / RL. Recovery is computed on pass@1. maj@16 is the accuracy of the most frequent final answer among the 16 samples, grouping answers that Math-Verify judges equivalent. Replace rate is the share of responses' tokens taken from the RL model, averaged per response.}
\label{tab:years}
\footnotesize
\begin{tabular}{@{}lc ccc c c@{}}
\toprule
Year & \# & pass@1 & pass@16 & maj@16 & Recovery & Replace rate \\
\midrule
AIME 2024 & 30 & 12.50 / 23.13 / 23.13 & 36.67 / 43.33 / 40.00 & 16.67 / 30.00 / 26.67 & 100.0 & 4.5 \\
AIME 2025 & 30 & \phantom{0}9.79 / 21.04 / 22.92 & 36.67 / 53.33 / 43.33 & 26.67 / 30.00 / 33.33 & \phantom{0}85.7 & 4.7 \\
AIME 2022 & 30 & \phantom{0}7.08 / 14.38 / 15.21 & 20.00 / 36.67 / 36.67 & 13.33 / 16.67 / 23.33 & \phantom{0}89.8 & 4.3 \\
AIME 2023 & 30 & 10.63 / 15.21 / 16.04 & 26.67 / 40.00 / 40.00 & 10.00 / 20.00 / 20.00 & \phantom{0}84.7 & 4.7 \\
\bottomrule
\end{tabular}
\end{table*}

\paragraph{Independent runs.}
An earlier, independent run of the full protocol on AIME 2024 and 2025 gives the same picture: \rlrep{} reaches 22.08\% and 20.00\% pass@1 against 23.13\% and 21.67\% for the RL model, recovering 90.1\% and 86.6\% of the gain. The larger run of \S\ref{sec:passk} (Table~\ref{tab:passk-main}) also reproduces the main result: from its $k=1$ values, \rlrep{} recovers 91.7\% of the RL gain on all 120 problems (88.0\% on AIME24+25 and 97.8\% on AIME22+23), against 90.5\% in the main run.

\paragraph{Run-to-run variation.}
Repeated evaluations of the same model are not identical. Across our Qwen3-4B runs, the base model's pass@1 on the 120 AIME problems ranges from approximately 8.2\% to 10.26\%, and the RL model's pass@1 on AIME 2025 ranges from 20.4\% to 23.3\% over six evaluations. We therefore interpret small differences between independent runs cautiously. The opening-only intervention improves pass@1 by 3.2 points and pass@16 by 5.0 points, while \textsc{rl-replace} improves pass@1 by 8.4 points. The hard rules lower pass@1 by 5.5--7.7 points. The roughly 1.5-point decreases from the soft rules are comparable to the variation between independent runs, so we report these conditions only as bringing no observed improvement.
\section{Conclusion}
\label{sec:conclusion}

We asked what RL adds to a base model on hard math problems by testing common explanations of the gain with interventions on the base model. Borrowing the RL model's token only at the positions where the two models' greedy tokens differ recovers most of the gain, whereas sharpening the base distribution, forcing the RL model's opening and inserting reflective tokens by fixed rules each recover at most a minority of it. Most of the gain comes after the opening: on top of the same opening, the RL model's later choices add 5.2 points, more than half of the RL gain, and using the base model's own greedy token at the same positions gives less than a third of this increment. These few decisions redirect the whole response, yet borrowing only them keeps much of the base model's diversity, and with the same number of attempts \rlrep{} solves more problems at large $k$ than either model or a response-level mixture matched in pass@1. The base model's own confidence already ranks the positions where RL changes the next-token distribution most (Appendix~\ref{app:predictor}); what remains hard is choosing the right token there.

\clearpage

\bibliographystyle{unsrtnat}
\bibliography{aaai2027_conference}

@article{wu2025leash,
  title   = {The Invisible Leash: Why {RLVR} May or May Not Escape Its Origin},
  author  = {Wu, Fang and Xuan, Weihao and Lu, Ximing and Liu, Mingjie and Dong, Yi and Harchaoui, Zaid and Choi, Yejin},
  journal = {arXiv preprint arXiv:2507.14843},
  year    = {2025}
}

@article{zhou2026cutting,
  title   = {Reasoning with Sampling: Cutting at Decision Points},
  author  = {Zhou, Felix and Mehrotra, Anay and Liu, Quanquan C.},
  journal = {arXiv preprint arXiv:2605.30327},
  year    = {2026},
  note    = {NeurIPS 2026}
}

@article{yang2026correctmass,
  title   = {More Correct Mass, Worse Answers: Why Power Sampling Can Fail and How to Fix It},
  author  = {Yang, Haohui and Sun, Jiaxing and Ma, Xiujun},
  journal = {arXiv preprint arXiv:2608.14420},
  year    = {2026}
}

@article{wang2025beyond,
  title   = {Beyond the 80/20 Rule: High-Entropy Minority Tokens Drive Effective Reinforcement Learning for {LLM} Reasoning},
  author  = {Wang, Shenzhi and Yu, Le and Gao, Chang and Zheng, Chujie and Liu, Shixuan and Lu, Rui and Dang, Kai and Chen, Xionghui and Yang, Jianxin and Zhang, Zhenru and Liu, Yuqiong and Yang, An and Zhao, Andrew and Yue, Yang and Song, Shiji and Yu, Bowen and Huang, Gao and Lin, Junyang},
  journal = {arXiv preprint arXiv:2506.01939},
  year    = {2025},
  note    = {NeurIPS 2025}
}

@article{huang2026direction,
  title   = {On the Direction of {RLVR} Updates for {LLM} Reasoning: Identification and Exploitation},
  author  = {Huang, Kexin and Meng, Haoming and Wu, Junkang and Lu, Jinda and Ma, Chiyu and Chen, Ziqian and Wang, Xue and Ding, Bolin and Wu, Jiancan and Wang, Xiang and He, Xiangnan and Wang, Guoyin and Zhou, Jingren},
  journal = {arXiv preprint arXiv:2603.22117},
  year    = {2026},
  note    = {ICLR 2026, as ``Beyond Magnitude: Leveraging Direction of RLVR Updates for LLM Reasoning''}
}

@article{shen2026restored,
  title   = {Reasoning Can Be Restored by Correcting a Few Decision Tokens},
  author  = {Shen, Changshuo and Sheng, Leheng and Chen, Yuxin and Zhang, An and Wang, Xiang},
  journal = {arXiv preprint arXiv:2605.16874},
  year    = {2026}
}

@article{wang2025wait,
  title   = {Wait, We Don't Need to ``Wait''! Removing Thinking Tokens Improves Reasoning Efficiency},
  author  = {Wang, Chenlong and Feng, Yuanning and Chen, Dongping and Chu, Zhaoyang and Krishna, Ranjay and Zhou, Tianyi},
  journal = {arXiv preprint arXiv:2506.08343},
  year    = {2025},
  note    = {Findings of EMNLP 2025}
}

@article{kang2025firsttry,
  title   = {First Try Matters: Revisiting the Role of Reflection in Reasoning Models},
  author  = {Kang, Liwei and Deng, Yue and Xiao, Yao and Mo, Zhanfeng and Lee, Wee Sun and Bing, Lidong},
  journal = {arXiv preprint arXiv:2510.08308},
  year    = {2025}
}

@article{deepseek2025r1,
  title   = {{DeepSeek-R1}: Incentivizing Reasoning Capability in {LLMs} via Reinforcement Learning},
  author  = {{DeepSeek-AI} and Guo, Daya and Yang, Dejian and Zhang, Haowei and Song, Junxiao and Wang, Peiyi and Zhu, Qihao and Xu, Runxin and Zhang, Ruoyu and Ma, Shirong and others},
  journal = {arXiv preprint arXiv:2501.12948},
  year    = {2025},
  note    = {Published in Nature, 2025, doi:10.1038/s41586-025-09422-z}
}

@article{zeng2025simplerl,
  title   = {{SimpleRL-Zoo}: Investigating and Taming Zero Reinforcement Learning for Open Base Models in the Wild},
  author  = {Zeng, Weihao and Huang, Yuzhen and Liu, Qian and Liu, Wei and He, Keqing and Ma, Zejun and He, Junxian},
  journal = {arXiv preprint arXiv:2503.18892},
  year    = {2025}
}

@article{zhao2025echo,
  title   = {Echo Chamber: {RL} Post-training Amplifies Behaviors Learned in Pretraining},
  author  = {Zhao, Rosie and Meterez, Alexandru and Kakade, Sham and Pehlevan, Cengiz and Jelassi, Samy and Malach, Eran},
  journal = {arXiv preprint arXiv:2504.07912},
  year    = {2025}
}

@article{shao2025spurious,
  title   = {Spurious Rewards: Rethinking Training Signals in {RLVR}},
  author  = {Shao, Rulin and Li, Shuyue Stella and Xin, Rui and Geng, Scott and Wang, Yiping and Oh, Sewoong and Du, Simon Shaolei and Lambert, Nathan and Min, Sewon and Krishna, Ranjay and Tsvetkov, Yulia and Hajishirzi, Hannaneh and Koh, Pang Wei and Zettlemoyer, Luke},
  journal = {arXiv preprint arXiv:2506.10947},
  year    = {2025}
}

@article{wang2025oneexample,
  title   = {Reinforcement Learning for Reasoning in Large Language Models with One Training Example},
  author  = {Wang, Yiping and Yang, Qing and Zeng, Zhiyuan and Ren, Liliang and Liu, Liyuan and Peng, Baolin and Cheng, Hao and He, Xuehai and Wang, Kuan and Gao, Jianfeng and Chen, Weizhu and Wang, Shuohang and Du, Simon Shaolei and Shen, Yelong},
  journal = {arXiv preprint arXiv:2504.20571},
  year    = {2025}
}

@article{matsutani2025squeezes,
  title   = {{RL} Squeezes, {SFT} Expands: A Comparative Study of Reasoning {LLMs}},
  author  = {Matsutani, Kohsei and Takashiro, Shota and Minegishi, Gouki and Kojima, Takeshi and Iwasawa, Yusuke and Matsuo, Yutaka},
  journal = {arXiv preprint arXiv:2509.21128},
  year    = {2025},
  note    = {ICLR 2026}
}

@article{qin2025decomposing,
  title   = {Decomposing Elements of Problem Solving: What ``Math'' Does {RL} Teach?},
  author  = {Qin, Tian and Park, Core Francisco and Kwun, Mujin and Walsman, Aaron and Malach, Eran and Anand, Nikhil and Tanaka, Hidenori and Alvarez-Melis, David},
  journal = {arXiv preprint arXiv:2505.22756},
  year    = {2025}
}

@article{liu2025prorl,
  title   = {{ProRL}: Prolonged Reinforcement Learning Expands Reasoning Boundaries in Large Language Models},
  author  = {Liu, Mingjie and Diao, Shizhe and Lu, Ximing and Hu, Jian and Dong, Xin and Choi, Yejin and Kautz, Jan and Dong, Yi},
  journal = {arXiv preprint arXiv:2505.24864},
  year    = {2025}
}

@article{li2025questa,
  title   = {{QuestA}: Expanding Reasoning Capacity in {LLMs} via Question Augmentation},
  author  = {Li, Jiazheng and Lin, Hongzhou and Lu, Hong and Wen, Kaiyue and Yang, Zaiwen and Gao, Jiaxuan and Wu, Yi and Zhang, Jingzhao},
  journal = {arXiv preprint arXiv:2507.13266},
  year    = {2025}
}

@article{yuan2025composing,
  title   = {From $f(x)$ and $g(x)$ to $f(g(x))$: {LLMs} Learn New Skills in {RL} by Composing Old Ones},
  author  = {Yuan, Lifan and Chen, Weize and Zhang, Yuchen and Cui, Ganqu and Wang, Hanbin and You, Ziming and Ding, Ning and Liu, Zhiyuan and Sun, Maosong and Peng, Hao},
  journal = {arXiv preprint arXiv:2509.25123},
  year    = {2025}
}

@article{gandhi2025cognitive,
  title   = {Cognitive Behaviors that Enable Self-Improving Reasoners, or, Four Habits of Highly Effective {STaRs}},
  author  = {Gandhi, Kanishk and Chakravarthy, Ayush and Singh, Anikait and Lile, Nathan and Goodman, Noah D.},
  journal = {arXiv preprint arXiv:2503.01307},
  year    = {2025}
}

@article{cui2025entropy,
  title   = {The Entropy Mechanism of Reinforcement Learning for Reasoning Language Models},
  author  = {Cui, Ganqu and Zhang, Yuchen and Chen, Jiacheng and Yuan, Lifan and Wang, Zhi and Zuo, Yuxin and Li, Haozhan and Fan, Yuchen and Chen, Huayu and Chen, Weize and Liu, Zhiyuan and Peng, Hao and Bai, Lei and Ouyang, Wanli and Cheng, Yu and Zhou, Bowen and Ding, Ning},
  journal = {arXiv preprint arXiv:2505.22617},
  year    = {2025}
}

@article{zhu2025negative,
  title   = {The Surprising Effectiveness of Negative Reinforcement in {LLM} Reasoning},
  author  = {Zhu, Xinyu and Xia, Mengzhou and Wei, Zhepei and Chen, Wei-Lin and Chen, Danqi and Meng, Yu},
  journal = {arXiv preprint arXiv:2506.01347},
  year    = {2025}
}

@article{wang2022selfconsistency,
  title   = {Self-Consistency Improves Chain of Thought Reasoning in Language Models},
  author  = {Wang, Xuezhi and Wei, Jason and Schuurmans, Dale and Le, Quoc and Chi, Ed and Narang, Sharan and Chowdhery, Aakanksha and Zhou, Denny},
  journal = {arXiv preprint arXiv:2203.11171},
  year    = {2022},
  note    = {ICLR 2023}
}

@article{wang2024cotdecoding,
  title   = {Chain-of-Thought Reasoning Without Prompting},
  author  = {Wang, Xuezhi and Zhou, Denny},
  journal = {arXiv preprint arXiv:2402.10200},
  year    = {2024}
}

@article{agarwal2025entropy,
  title   = {The Unreasonable Effectiveness of Entropy Minimization in {LLM} Reasoning},
  author  = {Agarwal, Shivam and Zhang, Zimin and Yuan, Lifan and Han, Jiawei and Peng, Hao},
  journal = {arXiv preprint arXiv:2505.15134},
  year    = {2025}
}

@article{talwar2026critical,
  title   = {A Critical Look at Power Sampling when Test Time Sharpening {LLMs}},
  author  = {Talwar, Abiel Malkani and Nemeth, Christopher and Moss, Henry},
  journal = {arXiv preprint arXiv:2610.13653},
  year    = {2026}
}

@article{liu2024proxy,
  title   = {Tuning Language Models by Proxy},
  author  = {Liu, Alisa and Han, Xiaochuang and Wang, Yizhong and Tsvetkov, Yulia and Choi, Yejin and Smith, Noah A.},
  journal = {arXiv preprint arXiv:2401.08565},
  year    = {2024}
}

@article{mitchell2023emulator,
  title   = {An Emulator for Fine-Tuning Large Language Models using Small Language Models},
  author  = {Mitchell, Eric and Rafailov, Rafael and Sharma, Archit and Finn, Chelsea and Manning, Christopher D.},
  journal = {arXiv preprint arXiv:2310.12962},
  year    = {2023}
}

@article{fei2024nudging,
  title   = {Nudging: Inference-time Alignment of {LLMs} via Guided Decoding},
  author  = {Fei, Yu and Razeghi, Yasaman and Singh, Sameer},
  journal = {arXiv preprint arXiv:2410.09300},
  year    = {2024}
}

@article{leviathan2022speculative,
  title   = {Fast Inference from Transformers via Speculative Decoding},
  author  = {Leviathan, Yaniv and Kalman, Matan and Matias, Yossi},
  journal = {arXiv preprint arXiv:2211.17192},
  year    = {2022}
}

@article{lin2023unlocking,
  title   = {The Unlocking Spell on Base {LLMs}: Rethinking Alignment via In-Context Learning},
  author  = {Lin, Bill Yuchen and Ravichander, Abhilasha and Lu, Ximing and Dziri, Nouha and Sclar, Melanie and Chandu, Khyathi and Bhagavatula, Chandra and Choi, Yejin},
  journal = {arXiv preprint arXiv:2312.01552},
  year    = {2023}
}

@article{zhou2023lima,
  title   = {{LIMA}: Less Is More for Alignment},
  author  = {Zhou, Chunting and Liu, Pengfei and Xu, Puxin and Iyer, Srini and Sun, Jiao and Mao, Yuning and Ma, Xuezhe and Efrat, Avia and Yu, Ping and Yu, Lili and Zhang, Susan and Ghosh, Gargi and Lewis, Mike and Zettlemoyer, Luke and Levy, Omer},
  journal = {arXiv preprint arXiv:2305.11206},
  year    = {2023}
}

@article{qi2024safety,
  title   = {Safety Alignment Should Be Made More Than Just a Few Tokens Deep},
  author  = {Qi, Xiangyu and Panda, Ashwinee and Lyu, Kaifeng and Ma, Xiao and Roy, Subhrajit and Beirami, Ahmad and Mittal, Prateek and Henderson, Peter},
  journal = {arXiv preprint arXiv:2406.05946},
  year    = {2024},
  note    = {ICLR 2025}
}

@article{mukherjee2025subnetworks,
  title   = {Reinforcement Learning Finetunes Small Subnetworks in Large Language Models},
  author  = {Mukherjee, Sagnik and Yuan, Lifan and Hakkani-Tur, Dilek and Peng, Hao},
  journal = {arXiv preprint arXiv:2505.11711},
  year    = {2025}
}

@article{liu2025understanding,
  title   = {Understanding {R1-Zero}-Like Training: A Critical Perspective},
  author  = {Liu, Zichen and Chen, Changyu and Li, Wenjun and Qi, Penghui and Pang, Tianyu and Du, Chao and Lee, Wee Sun and Lin, Min},
  journal = {arXiv preprint arXiv:2503.20783},
  year    = {2025}
}

@article{qian2025infopeaks,
  title   = {Demystifying Reasoning Dynamics with Mutual Information: Thinking Tokens are Information Peaks in {LLM} Reasoning},
  author  = {Qian, Chen and Liu, Dongrui and Wen, Haochen and Bai, Zhen and Liu, Yong and Shao, Jing},
  journal = {arXiv preprint arXiv:2506.02867},
  year    = {2025}
}

@article{muennighoff2025s1,
  title   = {s1: Simple test-time scaling},
  author  = {Muennighoff, Niklas and Yang, Zitong and Shi, Weijia and Li, Xiang Lisa and Fei-Fei, Li and Hajishirzi, Hannaneh and Zettlemoyer, Luke and Liang, Percy and Cand{\`e}s, Emmanuel and Hashimoto, Tatsunori},
  journal = {arXiv preprint arXiv:2501.19393},
  year    = {2025},
  note    = {EMNLP 2025}
}

@article{ghosal2025mirage,
  title   = {Does Thinking More always Help? Mirage of Test-Time Scaling in Reasoning Models},
  author  = {Ghosal, Soumya Suvra and Chakraborty, Souradip and Reddy, Avinash and Lu, Yifu and Wang, Mengdi and Manocha, Dinesh and Huang, Furong and Ghavamzadeh, Mohammad and Bedi, Amrit Singh},
  journal = {arXiv preprint arXiv:2506.04210},
  year    = {2025},
  note    = {NeurIPS 2025}
}

@article{chen2024overthinking,
  title   = {Do {NOT} Think That Much for 2+3=? On the Overthinking of o1-Like {LLMs}},
  author  = {Chen, Xingyu and Xu, Jiahao and Liang, Tian and He, Zhiwei and Pang, Jianhui and Yu, Dian and Song, Linfeng and Liu, Qiuzhi and Zhou, Mengfei and Zhang, Zhuosheng and Wang, Rui and Tu, Zhaopeng and Mi, Haitao and Yu, Dong},
  journal = {arXiv preprint arXiv:2412.21187},
  year    = {2024}
}

@article{chen2025seal,
  title   = {{SEAL}: Steerable Reasoning Calibration of Large Language Models for Free},
  author  = {Chen, Runjin and Zhang, Zhenyu and Hong, Junyuan and Kundu, Souvik and Wang, Zhangyang},
  journal = {arXiv preprint arXiv:2504.07986},
  year    = {2025}
}

@article{wang2025underthinking,
  title   = {Thoughts Are All Over the Place: On the Underthinking of o1-Like {LLMs}},
  author  = {Wang, Yue and Liu, Qiuzhi and Xu, Jiahao and Liang, Tian and Chen, Xingyu and He, Zhiwei and Song, Linfeng and Yu, Dian and Li, Juntao and Zhang, Zhuosheng and Wang, Rui and Tu, Zhaopeng and Mi, Haitao and Yu, Dong},
  journal = {arXiv preprint arXiv:2501.18585},
  year    = {2025}
}

@article{yang2025aha,
  title   = {Understanding Aha Moments: from External Observations to Internal Mechanisms},
  author  = {Yang, Shu and Wu, Junchao and Chen, Xin and Xiao, Yunze and Yang, Xinyi and Wong, Derek F. and Wang, Di},
  journal = {arXiv preprint arXiv:2504.02956},
  year    = {2025}
}

@article{lambert2024tulu3,
  title   = {{Tulu 3}: Pushing Frontiers in Open Language Model Post-Training},
  author  = {Lambert, Nathan and Morrison, Jacob and Pyatkin, Valentina and Huang, Shengyi and Ivison, Hamish and Brahman, Faeze and Miranda, Lester James V. and Liu, Alisa and Dziri, Nouha and Lyu, Shane and Gu, Yuling and Malik, Saumya and Graf, Victoria and Hwang, Jena D. and Yang, Jiangjiang and Le Bras, Ronan and Tafjord, Oyvind and Wilhelm, Chris and Soldaini, Luca and Smith, Noah A. and Wang, Yizhong and Dasigi, Pradeep and Hajishirzi, Hannaneh},
  journal = {arXiv preprint arXiv:2411.15124},
  year    = {2024}
}

@article{kimi2025k15,
  title   = {{Kimi k1.5}: Scaling Reinforcement Learning with {LLMs}},
  author  = {{Kimi Team} and Du, Angang and Gao, Bofei and Xing, Bowei and Jiang, Changjiu and Chen, Cheng and Li, Cheng and Xiao, Chenjun and Du, Chenzhuang and Liao, Chonghua and others},
  journal = {arXiv preprint arXiv:2501.12599},
  year    = {2025}
}

@article{wei2022cot,
  title   = {Chain-of-Thought Prompting Elicits Reasoning in Large Language Models},
  author  = {Wei, Jason and Wang, Xuezhi and Schuurmans, Dale and Bosma, Maarten and Ichter, Brian and Xia, Fei and Chi, Ed and Le, Quoc V. and Zhou, Denny},
  journal = {arXiv preprint arXiv:2201.11903},
  year    = {2022},
  note    = {NeurIPS 2022}
}

@article{li2023contrastive,
  title   = {Contrastive Decoding: Open-ended Text Generation as Optimization},
  author  = {Li, Xiang Lisa and Holtzman, Ari and Fried, Daniel and Liang, Percy and Eisner, Jason and Hashimoto, Tatsunori and Zettlemoyer, Luke and Lewis, Mike},
  journal = {arXiv preprint arXiv:2210.15097},
  year    = {2023},
  note    = {ACL 2023}
}

@article{chen2025passk,
  title   = {Pass@k Training for Adaptively Balancing Exploration and Exploitation of Large Reasoning Models},
  author  = {Chen, Zhipeng and Qin, Xiaobo and Wu, Youbin and Ling, Yue and Ye, Qinghao and Zhao, Wayne Xin and Shi, Guang},
  journal = {arXiv preprint arXiv:2508.10751},
  year    = {2025}
}

@article{gema2025inverse,
  title   = {Inverse Scaling in Test-Time Compute},
  author  = {Gema, Aryo Pradipta and H{\"a}gele, Alexander and Chen, Runjin and Arditi, Andy and Goldman-Wetzler, Jacob and Fraser-Taliente, Kit and Sleight, Henry and Petrini, Linda and Michael, Julian and Alex, Beatrice and Minervini, Pasquale and Chen, Yanda and Benton, Joe and Perez, Ethan},
  journal = {arXiv preprint arXiv:2507.14417},
  year    = {2025}
}

@article{karan2026sampling,
  title   = {Reasoning with Sampling: Your Base Model is Smarter Than You Think},
  author  = {Karan, Aayush and Du, Yilun},
  journal = {arXiv preprint arXiv:2510.14901},
  year    = {2025},
  note    = {ICLR 2026}
}

@article{wang2026cue,
  title   = {Base Models Can Reason By Taking a Cue From Training Data},
  author  = {Wang, Sophie L. and Dravid, Amil and Shao, Rulin and Farhat, Kevin and Min, Sewon and Efros, Alexei A.},
  journal = {arXiv preprint arXiv:2610.06851},
  year    = {2026}
}

@article{meng2026sparse,
  title   = {Sparse but Critical: A Token-Level Analysis of Distributional Shifts in {RLVR} Fine-Tuning of {LLMs}},
  author  = {Meng, Haoming and Huang, Kexin and Wei, Shaohang and Ma, Chiyu and Yang, Shuo and Wang, Xue and Wang, Guoyin and Ding, Bolin and Zhou, Jingren},
  journal = {arXiv preprint arXiv:2603.22446},
  year    = {2026},
  note    = {ICLR 2026}
}

@article{akgul2026sparse,
  title   = {Rethinking {RL} for {LLM} Reasoning: It's Sparse Policy Selection, Not Capability Learning},
  author  = {Akg{\"u}l, {\"O}mer Faruk and Kannan, Rajgopal and Neiswanger, Willie and Prasanna, Viktor},
  journal = {arXiv preprint arXiv:2605.06241},
  year    = {2026}
}

@article{venhoff2026when,
  title   = {Base Models Know How to Reason, Thinking Models Learn When},
  author  = {Venhoff, Constantin and Arcuschin, Iv{\'a}n and Torr, Philip and Conmy, Arthur and Nanda, Neel},
  journal = {arXiv preprint arXiv:2510.07364},
  year    = {2025},
  note    = {ICML 2026}
}

@article{yue2025doesrl,
  title   = {Does Reinforcement Learning Really Incentivize Reasoning Capacity in {LLMs} Beyond the Base Model?},
  author  = {Yue, Yang and Chen, Zhiqi and Lu, Rui and Zhao, Andrew and Wang, Zhaokai and Yue, Yang and Song, Shiji and Huang, Gao},
  journal = {arXiv preprint arXiv:2504.13837},
  year    = {2025},
  note    = {NeurIPS 2025}
}

@article{yu2025dapo,
  title   = {{DAPO}: An Open-Source {LLM} Reinforcement Learning System at Scale},
  author  = {Yu, Qiying and others},
  journal = {arXiv preprint arXiv:2503.14476},
  year    = {2025}
}

@article{shao2024deepseekmath,
  title   = {{DeepSeekMath}: Pushing the Limits of Mathematical Reasoning in Open Language Models},
  author  = {Shao, Zhihong and others},
  journal = {arXiv preprint arXiv:2402.03300},
  year    = {2024}
}

@article{chen2021codex,
  title   = {Evaluating Large Language Models Trained on Code},
  author  = {Chen, Mark and others},
  journal = {arXiv preprint arXiv:2107.03374},
  year    = {2021}
}

@article{yang2025qwen3,
  title   = {Qwen3 Technical Report},
  author  = {Yang, An and others},
  journal = {arXiv preprint arXiv:2505.09388},
  year    = {2025}
}

@article{grattafiori2024llama3,
  title   = {The {Llama} 3 Herd of Models},
  author  = {Grattafiori, Aaron and others},
  journal = {arXiv preprint arXiv:2407.21783},
  year    = {2024}
}

@article{sheng2024hybridflow,
  title   = {{HybridFlow}: A Flexible and Efficient {RLHF} Framework},
  author  = {Sheng, Guangming and Zhang, Chi and Ye, Zilingfeng and Wu, Xibin and Zhang, Wang and Zhang, Ru and Peng, Yanghua and Lin, Haibin and Wu, Chuan},
  journal = {arXiv preprint arXiv:2409.19256},
  year    = {2024}
}

@article{lightman2023verify,
  title   = {Let's Verify Step by Step},
  author  = {Lightman, Hunter and Kosaraju, Vineet and Burda, Yura and Edwards, Harri and Baker, Bowen and Lee, Teddy and Leike, Jan and Schulman, John and Sutskever, Ilya and Cobbe, Karl},
  journal = {arXiv preprint arXiv:2305.20050},
  year    = {2023}
}

@inproceedings{kwon2023vllm,
  title     = {Efficient Memory Management for Large Language Model Serving with {PagedAttention}},
  author    = {Kwon, Woosuk and Li, Zhuohan and Zhuang, Siyuan and Sheng, Ying and Zheng, Lianmin and Yu, Cody Hao and Gonzalez, Joseph E. and Zhang, Hao and Stoica, Ion},
  booktitle = {Proceedings of the 29th Symposium on Operating Systems Principles},
  year      = {2023}
}

@misc{mathverify2025,
  title        = {{Math-Verify}: A Robust Mathematical Expression Evaluation System},
  author       = {Kydl{\'i}{\v{c}}ek, Hynek and {Hugging Face}},
  howpublished = {\url{https://github.com/huggingface/Math-Verify}},
  year         = {2025}
}

\clearpage
\appendix

\section{Implementation Details}
\label{app:impl}

\subsection{The \rlrep{} decoder}
\label{app:impl-rlrep}

\rlrep{} generates one response at a time with the base model and the RL model reading the same prefix. Starting from the prompt of Appendix~\ref{app:prompts}, each step proceeds as follows.
\begin{enumerate}
  \item Both models score the current prefix $y_{<t}$, giving $p_B(\cdot\mid y_{<t})$ and $p_R(\cdot\mid y_{<t})$.
  \item The two greedy tokens $b_t=\arg\max p_B(\cdot\mid y_{<t})$ and $r_t=\arg\max p_R(\cdot\mid y_{<t})$ are compared.
  \item If $b_t\neq r_t$, the step is a disagreement position and $r_t$ is emitted. Otherwise $y_t$ is sampled from $p_B$ at temperature 0.6 with top-$p$ 0.95.
  \item The emitted token is appended to the prefix of both models, and generation stops at the end-of-sequence token or after 8{,}192 tokens.
\end{enumerate}
Because each RL model is trained from its own base model, the two share a tokenizer and the comparison is made on token ids. Two different ids can decode to the same text: in the 38 responses analysed in Appendix~\ref{app:replacements}, 9{,}472 tokens were emitted by the RL model, of which 9{,}099 (96.1\%) change the text and 373 differ only in tokenization. The replace rate of a response is the number of tokens emitted by the RL model divided by the response length; the per-token rate pools all tokens of a run and gives more weight to long responses, which is why it is slightly lower (4.0\%) than the per-response average (4.5\%).

The rule involves no threshold. The RL model is consulted only through its greedy token, so \rlrep{} never samples from the RL model, and at the positions where the two greedy tokens agree the output is distributed exactly as standard base sampling.

\subsection{Settings of each experiment}
\label{app:impl-settings}

Table~\ref{tab:settings} summarizes the experimental settings. All evaluations use the prompt of Appendix~B and a limit of 8,192 generated tokens. Unless otherwise specified, standard sampling uses temperature 0.6 and top-$p$ 0.95. In \textsc{rl-replace}, these settings apply to base-model sampling at agreement positions; disagreement positions use the RL model's greedy token. Greedy controls, power sampling, and temperature ablations follow the settings specified in their respective sections.

\begin{table*}[t]
\centering
\caption{Settings of the experiments in the paper. ``Samples'' is per problem.}
\label{tab:settings}
\footnotesize
\setlength{\tabcolsep}{4pt}
\begin{tabular}{@{}llp{2.0cm}p{2.2cm}p{1.9cm}p{3.4cm}@{}}
\toprule
Experiment & Where & Benchmark & Models & Samples & Decoding \\
\midrule
Power sampling (Test~0) & Table~\ref{tab:mcmc} & MATH-500, AIME & 3 pairs / Qwen3-4B/8B & 8 seeds; greedy once & MCMC, $\alpha=4$, 16 blocks $\times$ 10 steps \\
Gate-MCMC & \S\ref{sec:sampling} & AIME (120) & Qwen3-4B & 16 & MCMC where gate $\geq 0.8567$ \\
\rlrep{} (Test~1) & Table~\ref{tab:main} & AIME (120), MATH-500 & Qwen3-4B & 16; 64 on MATH-500 & Appendix~\ref{app:impl-rlrep} \\
Opening ablation (Test~2) & Table~\ref{tab:firsttoken} & AIME (120) & Qwen3-4B & 16 & Prefilled ``Alright'' / ``Alright, I'' \\
Greedy control & Table~\ref{tab:firsttoken} & AIME (120) & Qwen3-4B & 16 & ``Alright'', base greedy at disagreements \\
Rules (Test~3) & Table~\ref{tab:rules} & AIME (120) & Qwen3-4B & 16 & Hard insertion or logit bias $+2$ / $+4$ \\
Large $k$& \shortstack[l]{ Table~\ref{tab:passk-main} \\ Figure~\ref{fig:passk} (4B) } & AIME (120) & Qwen3-4B/8B & 512; 256 for \rlrep{} & Standard sampling / \rlrep{} \\
Other model pairs & \S\ref{sec:exp-models} & MATH-500 & Qwen3-8B, Llama-3.2-3B & as in Table~\ref{tab:mcmc} & Sampling, MCMC, \rlrep{} \\
High-KL predictor & Appendix~\ref{app:predictor} & AIME24+25 & Qwen3-4B & 16 base responses & Scored by both models \\
\bottomrule
\end{tabular}
\end{table*}

\subsection{RL training}
\label{app:impl-rl}

Table~\ref{tab:rl-train} lists the training configuration shared by the three RL models. Hyperparameters not listed are the defaults of verl's GRPO recipe.

\begin{table}[h]
\centering
\caption{RL training configuration, shared by all three model pairs.}
\label{tab:rl-train}
\small
\begin{tabular}{@{}ll@{}}
\toprule
Setting & Value \\
\midrule
Algorithm & GRPO~\citep{shao2024deepseekmath} \\
Framework & verl~\citep{sheng2024hybridflow} \\
Initialization & Base model, no supervised fine-tuning \\
Training data & DAPO-Math-17k~\citep{yu2025dapo} \\
Prompt format & DAPO-Math-17k format (Appendix~\ref{app:prompts}) \\
Learning rate & $10^{-6}$, constant \\
Reward & $+1$ for a correct final answer, $-1$ otherwise \\
Prompts per step & 128 \\
Rollouts per prompt & 8 \\
Checkpoint selection & Best accuracy on a DAPO-Math-17k validation split \\
Selected step & 460 (Qwen3-4B), 320 (Qwen3-8B), 500 (Llama-3.2-3B) \\
\bottomrule
\end{tabular}
\end{table}

\paragraph{Training dynamics.}
The Qwen3-4B run follows the usual pattern of RL applied directly to a base model. Training accuracy rises from 0.071 at the first step to 0.413 at step 50 and reaches a plateau of 0.50--0.54 after about 150 steps (0.503 at step 460, the checkpoint we use). Over the same period the median rollout length grows from 521 to 3{,}917 tokens, and the fraction of rollouts without a valid final answer falls from 0.408 to 0.027. Training runs on 8 GPUs with FSDP, and every step draws 128 prompts with 8 rollouts each, without resampling.

\subsection{Compute}
\label{app:impl-compute}

All generation uses vLLM~\citep{kwon2023vllm} on NVIDIA H800 GPUs. \rlrep{} runs a forward pass of both models for every generated token, with the two models served by separate vLLM engines on the same GPU. One AIME year with 16 samples per problem takes 7.3--7.9 GPU-hours, MATH-500 about 55 GPU-hours, and the large-$k$ run with 256 samples per problem on all four years about 1{,}012 GPU-hours.

\section{Prompt Templates}
\label{app:prompts}

\paragraph{Evaluation.}
Every model and intervention in this paper is evaluated with the same two messages. For Qwen3 models, applying the chat template with thinking disabled gives the following input, where \texttt{\{question\}} is the problem statement:

\begin{small}
\begin{verbatim}
<|im_start|>system
You are a helpful assistant.<|im_end|>
<|im_start|>user
{question}
Please reason step by step, and put your final answer
within \boxed{}.<|im_end|>
<|im_start|>assistant
<think>

</think>

\end{verbatim}
\end{small}

\noindent Line breaks inside the instruction texts, here and below, are for display only. Llama-3.2-3B receives the same system and user messages through its own chat template. Interventions that prefill the response (Test~2) append the forced text directly after this input.

\paragraph{RL training.}
RL rollouts use the prompt format distributed with DAPO-Math-17k~\citep{yu2025dapo}, with a single user turn:

\begin{small}
\begin{verbatim}
Solve the following math problem step by step. The last
line of your response should be of the form Answer: $Answer
(without quotes) where $Answer is the answer to the problem.

{question}

Remember to put your answer on its own line after "Answer:".
\end{verbatim}
\end{small}

\section{Metrics and Confidence Intervals}
\label{app:stats}

\paragraph{pass@$k$.}
For a problem with $n$ samples of which $c$ are correct, we use the unbiased estimator of \citet{chen2021codex},
\begin{equation}
  \text{pass@}k = 1-\binom{n-c}{k}\Big/\binom{n}{k},
\end{equation}
and average it over problems. pass@1 is then the mean accuracy over all samples, and when $k=n$ the estimator reduces to the fraction of problems solved at least once. The value for all 120 AIME problems is the mean of the two groups of 60, which equals the mean over problems.

\paragraph{Recovery rate.}
The recovery rate of \S\ref{sec:setup} is a ratio of two pass@1 differences measured on the same problems. Its interval is obtained with a problem-level bootstrap: we draw 10{,}000 resamples of the problems with replacement, within each group of 60 AIME problems (or within MATH-500), recompute both differences on every resample, and take the 95\% interval of the resulting ratios. Because the denominator, the RL gain, also varies across resamples, the intervals are wide and can extend above 100\% even when the point estimate is below it.

\paragraph{Differences between conditions.}
When two conditions are compared, both are evaluated on the same resampled problems, so the interval reflects only the variation of their difference across problems. Table~\ref{tab:cis} collects the intervals used in the paper.

\begin{table}[h]
\centering
\caption{Point estimates and 95\% bootstrap intervals (problem-level, paired across conditions; 10{,}000 resamples, and 5{,}000 for the difference to base + ``Alright''). AIME results use the 120 problems unless a group is named.}
\label{tab:cis}
\small
\setlength{\tabcolsep}{4pt}
\begin{tabular}{@{}llr@{}}
\toprule
Quantity & Metric & Estimate [95\% CI] \\
\midrule
\multicolumn{3}{@{}l}{Recovery rate of \rlrep{} (\%)} \\
\quad AIME24+25 & pass@1 & 92.1 [75.6, 121.2] \\
\quad AIME22+23 & pass@1 & 87.7 [63.8, 120.9] \\
\quad All AIME & pass@1 & 90.5 [77.0, 108.6] \\
\quad MATH-500 & pass@1 & 106.6 [95.2, 120.4] \\
\quad AIME24+25 minus AIME22+23 & pass@1 & 4.4 [$-$32, +41] \\
\midrule
\multicolumn{3}{@{}l}{Differences (points)} \\
\quad \rlrep{} $-$ (base + ``Alright'') & pass@1 & +5.21 [+3.12, +7.55] \\
\quad \rlrep{} $-$ RL & pass@256 & +6.10 [+0.82, +11.02] \\
\quad \rlrep{} $-$ base & pass@256 & +5.86 [$-$0.33, +10.71] \\
\quad \rlrep{} $-$ response mixture & pass@256 & +6.05 [+0.93, +10.98] \\
\bottomrule
\end{tabular}
\end{table}

\paragraph{Per-problem comparison.}
The pass@1 differences in Table~\ref{tab:main} are spread over many problems. On AIME24+25, \rlrep{} has a higher per-problem accuracy than the base model on 26 problems and a lower one on 4; against the RL model it is higher on 11 and lower on 13. On AIME22+23 the corresponding counts are 21 and 0 against the base model and 7 and 9 against the RL model. The numbers of problems solved at least once in 16 samples are 22, 29 and 25 on AIME24+25 and 14, 23 and 23 on AIME22+23 for the base model, \rlrep{} and the RL model, as reflected in the pass@16 columns of Table~\ref{tab:main}.

\paragraph{Majority voting.}
maj@16 takes the most frequent final answer among the 16 samples of a problem, grouping answers that Math-Verify judges equivalent, and scores it against the reference answer.

\section{Power Sampling Details}
\label{app:mcmc}

We follow the power-sampling MCMC of \citet{karan2026sampling}: the target distribution is the sequence-level power distribution $p_{\text{base}}^{\alpha}$ with $\alpha=4$, which differs from token-level sampling at temperature $1/\alpha$. Each response is generated in 16 blocks, and each block is refined with 10 MCMC steps. Standard sampling (at temperature 0.6) and MCMC use 8 seeds (0--7), and greedy decoding is run once. All runs share the 8{,}192-token generation limit of the main experiments.

\begin{table}[h]
\centering
\caption{Power sampling on AIME with Qwen3-4B-Base, pass@1 (\%).}
\label{tab:mcmc-aime}
\small
\setlength{\tabcolsep}{5pt}
\begin{tabular}{@{}lccc@{}}
\toprule
Problems & Greedy & Standard sampling & MCMC \\
\midrule
AIME 2025 (30) & 13.33 & \phantom{0}9.58 & 13.75 \\
AIME 2022--2024 (90) & 11.11 & 10.00 & 10.00 \\
\bottomrule
\end{tabular}
\end{table}

On AIME 2022--2024, MCMC matches standard sampling and stays below greedy decoding. On AIME 2025 it matches greedy decoding. On AIME, MCMC also degrades the outputs: the fraction of responses without a final answer rises from 12.2\% to 22.4\%, and the fraction that ends in repetition rises from 6.7\% to 13.1\%.

\paragraph{Gate-MCMC.}
The gated variant in \S\ref{sec:sampling} applies MCMC only at positions where the gradient-boosted gate of Appendix~\ref{app:predictor} scores at least 0.8567, which uses only the six base-model features listed there. On the 120 AIME problems with 16 samples per problem it obtains 9.84 / 24.17 pass@1 / pass@16, against 10.00 / 30.00 for standard base sampling.

\section{Locating High-KL Positions from the Base Model}
\label{app:predictor}

Section~\ref{sec:sparse} shows that the RL gain is concentrated at the few disagreement positions, where the two models' greedy tokens differ. Here we ask whether positions where the two models diverge can be located using only the base model, without running the RL model. This predictor is used in two places in the main text: the gate in gate-MCMC (\S\ref{sec:sampling}) and the trigger statistics behind the hard rules (\S\ref{sec:rules}).

\paragraph{Data and features.}
The base model generates 16 responses to each of the 30 AIME 2024 and 30 AIME 2025 problems at temperature 0.6, and the RL model scores every prefix of these responses. The two trained predictors in Table~\ref{tab:predictor}, logistic regression and the gate, are fit on responses to DAPO-Math-17k problems, the data used for RL training, and all AUROC values are computed on the AIME responses. For each token we record six features computed from the base model alone: the entropy of its next-token distribution, the probability of its top candidate, the margin between its top two candidates, the log-probability of the token it actually sampled (chosen log-prob), the rank of that token, and the relative position in the response.

\paragraph{Labels.}
We evaluate two labels for the positions where the two models differ. The \emph{KL} label marks the 10\% of positions with the largest KL divergence between the RL and base next-token distributions (the high-KL positions of the main text); it depends only on the two distributions, not on the token the base model sampled. The \emph{mismatch} label marks positions where the RL model's top-1 token differs from the token the base model sampled. This is not the criterion of \rlrep{}, which compares the two models' top-1 tokens: the mismatch label is also positive wherever the base model samples a token other than its own top-1, even if the two distributions are identical.

\begin{table}[h]
\centering
\caption{AUROC of base-only features for locating the positions where the two models differ (AIME24 / AIME25). The gate is a gradient-boosted classifier trained on all six features.}
\label{tab:predictor}
\footnotesize
\setlength{\tabcolsep}{4pt}
\begin{tabular}{@{}lcc@{}}
\toprule
 & Mismatch label & KL label (top 10\%) \\
\midrule
Chosen log-prob & 0.993 / 0.993 & 0.944 / 0.960 \\
Entropy & -- & 0.945 / 0.960 \\
Top-1 probability & -- & 0.945 / 0.963 \\
Margin & -- & 0.943 / 0.965 \\
Relative position & -- & 0.547 / 0.572 \\
Token rank & -- & 0.531 / 0.535 \\
\midrule
Logistic regression (all features) & -- & 0.946 / 0.958 \\
Gate (gradient boosting) & -- & 0.969 / 0.980 \\
\bottomrule
\end{tabular}
\end{table}

\paragraph{Results.}
High-KL positions are highly predictable from the base model's own confidence (Table~\ref{tab:predictor}). Under the KL label, every confidence feature reaches an AUROC of 0.94--0.96, and the gate trained on all features reaches 0.969 / 0.980, with a precision of 0.54 / 0.62 at 90\% recall. Position and token rank carry almost no information (0.53--0.57): whether a position diverges depends on how unsure the base model is there, not on where it falls in the response. Under the mismatch label, the chosen log-prob alone reaches 0.993 on both years. This number overstates how well the positions used by \rlrep{} can be located: a low chosen log-prob means the base model sampled a token other than its own top-1, which makes the mismatch label positive almost by construction. Our conclusions therefore rest on the KL label.

The numbers depend on the label. With a label defined as the 10\% of positions with the lowest $\log p_{\text{RL}} - \log p_{\text{base}}$, the best single feature drops to 0.63 on a separate set of 1{,}324 MATH problems; relabelling the same data with the KL label raises entropy to 0.95. Because most tokens have a positive log-probability difference, the lowest 10\% is a noisy label. We therefore use the KL label, which depends only on the two models' next-token distributions.

\paragraph{Implication.}
The base model knows \emph{where} it is uncertain: its confidence ranks the positions at which the RL model changes the next-token distribution most, with an AUROC of up to 0.98. What the base-only strategies we tested do not supply is \emph{which} token to use there. Gate-MCMC resamples at these positions and does not help (\S\ref{sec:sampling}), and rules that insert fixed tokens at similar positions hurt (\S\ref{sec:rules}). Locating these positions is comparatively easy; making the decisions is what the RL model contributes.

\section{What and Where the RL Model Replaces}
\label{app:replacements}

This appendix gives the details behind \S\ref{sec:sparse} and \S\ref{sec:rules}. We first describe the replacements in all 1{,}920 \rlrep{} responses of the main run, and then look closely at the 19 AIME problems on which the base model fails all 16 samples while \rlrep{} solves at least one, and the 38 successful \rlrep{} responses on them.

\subsection{All responses}
\label{app:rep-all}

\paragraph{How many tokens are replaced.}
The per-response replace rate is 4.51\% on AIME 2024, 4.69\% on AIME 2025, 4.31\% on AIME 2022 and 4.67\% on AIME 2023; weighted by tokens it is 3.84\%, 4.18\%, 3.99\% and 4.00\%. A response contains about 246--261 replacements. Correct responses have a lower replace rate than incorrect ones (3.96\% against 4.67\%). Although only about 4\% of the tokens come from the RL model, the responses take on its length: \rlrep{} responses average 6{,}217 tokens, close to the RL model's 6{,}504 and nearly twice the base model's 3{,}248.

\paragraph{Where.}
All 1{,}920 responses are replaced at the first position, always by ``Alright''. The base token it replaces is mostly ``To'' (51.5\%) or ``Assistant'' (42.8\%), and the most frequent replacement at position~2 is ``let''$\to$``I'' (976 times). On MATH-500, 94.8\% of the responses are replaced at the first position. After the opening, replacements are densest early in the response. Splitting each response into ten equal parts, the replace rate in each part is 6.0, 5.1, 4.4, 4.0, 3.7, 3.4, 3.3, 3.1, 3.2 and 3.8\%, so the first tenth has nearly twice the density of the second half. The 10th, 25th, 50th, 75th and 90th percentiles of the relative position of a replacement are 0.06, 0.17, 0.40, 0.70 and 0.89.

\paragraph{What.}
The replacements form a long tail: the main run contains 83{,}710 distinct base$\to$RL token pairs and 6{,}376 distinct RL tokens, and the 100 most frequent pairs account for only 16.9\% of all replacements. The most frequent pairs are ``.\textbackslash n\textbackslash n''$\to$``.'' (4{,}009), ``the''$\to$``\texttt{\textbackslash(}'' (3{,}781), ``the''$\to$``we'' (2{,}676), ``Perhaps''$\to$``Maybe'' (1{,}309), ``I''$\to$``we'' (1{,}213) and ``So''$\to$``Wait'' (872). The tokens the RL model inserts most often are ``the'' (18{,}123), ``\texttt{\textbackslash(}'' (17{,}768), ``we'' (10{,}339), ``\#\#\#'' (9{,}928) and ``But'' (5{,}361). Reflective words become far more frequent: ``Wait'' appears 0.30 times per base response and 5.74 times per \rlrep{} response, and ``Re'' is inserted by the RL model 2{,}204 times but replaced only 288 times.

\subsection{Successful responses on problems missed by all 16 base samples}
\label{app:rep-hard}

\begin{figure}[H]
    \centering
    \includegraphics[width=1\linewidth]{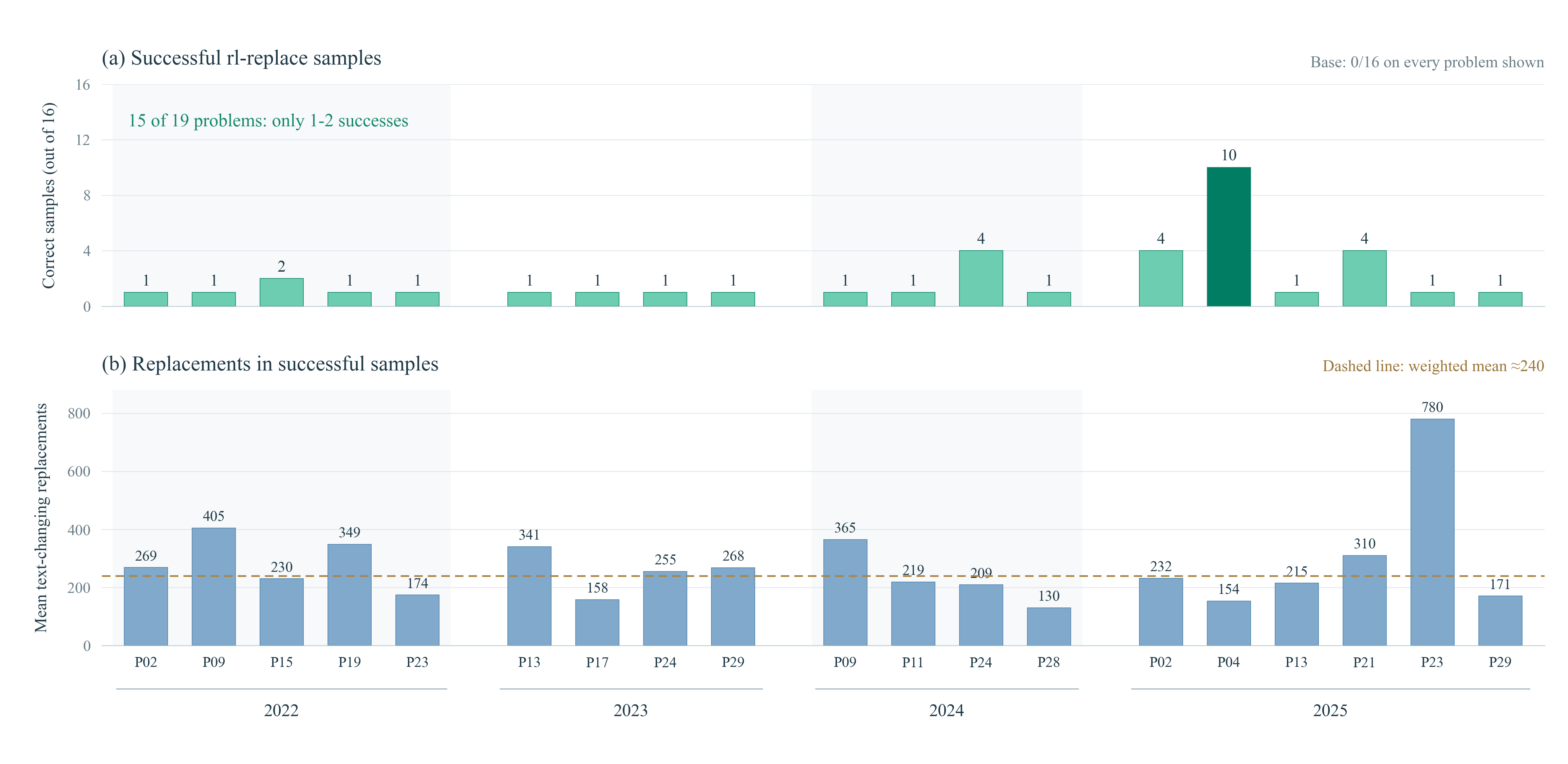}
    \caption{Results for the 19 problems on which the base model fails all 16 samples and \rlrep{} solves at least one, grouped by year. (a) Number of successful \rlrep{} samples out of 16. (b) Mean number of text-changing replacements per successful response. The dashed line marks the overall mean of approximately 240, weighted by the number of successful samples for each problem.}
    \label{fig:problem_result}
\end{figure}

\paragraph{Problems.}
Figure~\ref{fig:problem_result} summarizes the 19 problems on which the base model fails all 16 samples but \rlrep{} solves at least one. Panel~(a) shows the number of successful \rlrep{} samples for each problem. Fifteen of the 19 problems are solved by only one or two samples, while AIME 2025 Problem~4 stands out with 10 successes. Panel~(b) shows the mean number of text-changing replacements in the successful responses for each problem. Across all 38 successful responses, the mean is about 240 replacements; this overall mean weights each problem by its number of successful samples.

\paragraph{Position in the response.}
Table~\ref{tab:rep-position} splits the replacements by their relative position in the response. They are densest in the first fifth of the response and sparsest between 50\% and 80\%, with a slight rise in the last tenth.

\begin{table}[h]
\centering
\caption{Share of the 9{,}099 replacements by relative position in the response (\%).}
\label{tab:rep-position}
\small
\setlength{\tabcolsep}{3pt}
\begin{tabular}{@{}lcccccccccc@{}}
\toprule
Position (\%) & 0--10 & 10--20 & 20--30 & 30--40 & 40--50 & 50--60 & 60--70 & 70--80 & 80--90 & 90--100 \\
\midrule
Share & 14.0 & 12.1 & 11.6 & 10.0 & 10.7 & 8.2 & 7.8 & 8.0 & 8.5 & 9.1 \\
\bottomrule
\end{tabular}
\end{table}

\paragraph{Direction of the replacements.}
Table~\ref{tab:categories} in the main text merges categories that occur in both directions. Table~\ref{tab:rep-direction} separates them. The RL model moves text into \LaTeX{} more often than out of it (607 against 467) and introduces a first-person subject more often than it removes one (292 against 142), but both directions are common, so it has no fixed preference for either style.

\begin{table}[h]
\centering
\caption{Replacement categories with direction. ``To X'' means the RL token is of type X where the base model's token was not; ``from X'' is the reverse.}
\label{tab:rep-direction}
\small
\begin{tabular}{@{}llrr@{}}
\toprule
Category & Example & Count & Share (\%) \\
\midrule
Other content words & & 6{,}794 & 74.7 \\
To \LaTeX{} & the $\to$ \texttt{\textbackslash(} & 607 & 6.7 \\
From \LaTeX{} & \texttt{\textbackslash(} $\to$ : & 467 & 5.1 \\
To a first-person subject & the $\to$ we, let $\to$ I & 292 & 3.2 \\
Numbers & 2 $\to$ 4 & 202 & 2.2 \\
Opening and pause words & To $\to$ Alright, So $\to$ Wait & 190 & 2.1 \\
Markdown headers & Let $\to$ \#\#\# & 185 & 2.0 \\
Punctuation & , $\to$ . & 177 & 1.9 \\
From a first-person subject & & 142 & 1.6 \\
``Re-'' prefix & Correct $\to$ Re & 43 & 0.5 \\
\midrule
Total & & 9{,}099 & 100.0 \\
\bottomrule
\end{tabular}
\end{table}

\paragraph{Frequent replacement pairs.}
Table~\ref{tab:rep-pairs} lists the most frequent pairs by function. At the first position every one of the 38 responses replaces the base model's opening with ``Alright'', and in 25 of them the token at position~2 is also replaced, turning ``let'' into ``I'' and giving ``Alright, I have this problem''. Some pairs only change style, such as ``perhaps'' $\to$ ``maybe'' (18 times). Number replacements have no fixed direction; the most frequent pairs are 2$\to$1 (16), 2$\to$4 (12), 2$\to$6 (11), 1$\to$2 (10) and 0$\to$1 (10).

\begin{table}[h]
\centering
\caption{Frequent replacement pairs (base token $\to$ RL token) in the 38 responses, grouped by function.}
\label{tab:rep-pairs}
\small
\begin{tabular}{@{}llr@{}}
\toprule
Function & Pair & Count \\
\midrule
\multirow{3}{*}{Opening (position 0)} & To $\to$ Alright & 21 \\
 & Assistant $\to$ Alright & 9 \\
 & user $\to$ Alright & 8 \\
\midrule
\multirow{3}{*}{Pause before going on} & \#\#\# $\to$ Wait & 30 \\
 & So $\to$ Wait & 20 \\
 & But $\to$ Wait & 8 \\
\midrule
Question a conclusion & So $\to$ But & 20 \\
New section & Let, But $\to$ \#\#\# & 24 \\
\midrule
\multirow{3}{*}{Notation} & the $\to$ \texttt{\textbackslash(} & 59 \\
 & \texttt{\textbackslash(} $\to$ : & 43 \\
 & (none) $\to$ \texttt{\textbackslash(} & 32 \\
\midrule
Style & perhaps $\to$ maybe & 18 \\
\bottomrule
\end{tabular}
\end{table}

\paragraph{Excerpts.}
The following excerpts from successful \rlrep{} responses show the replaced token in brackets. Where the base model's greedy token opens a new section after a result, the RL model's token is a pause, followed by a check:
\begin{quote}\small
CE = 104\\
\textbf{[Wait]}, but let's check the calculation more precisely.
\end{quote}
Here the base model's greedy token is ``\#\#\#'', the start of a new header. Similarly, after a total is computed:
\begin{quote}\small
Total = 404 + 405 = 809.\\
\textbf{[Wait]}, but we need to ensure that we are not double-counting.
\end{quote}
The text that follows such a pause often states the problem explicitly, for example ``Wait, that's not correct. Let's think again.'' or ``Wait, this is inconsistent. Maybe I need to re-examine the small cases.'', and a pause is sometimes followed by a new header such as ``\#\#\# Re-examining the Problem''. Replacing ``So'' by ``But'' turns a conclusion into a check, as in ``\textbf{[But]} Alice can also remove 4 tokens, which is impossible. So, Bob cannot win.'' Number replacements fix values that the problem's constraints rule out. In a game where a player may remove one or four tokens, the base model writes 2 at the bracketed positions, and the RL model's tokens give
\begin{quote}\small
\ldots or she can try to remove \textbf{[4]} tokens, but she can't because there's only \textbf{[1]} token.
\end{quote}
In a coordinate-geometry problem, the base model's ``DA = 0'' becomes ``DA = \textbf{[6]}, so point D = (0, 6)''. In each case the replaced token depends on what has been derived so far, which is what the fixed rules of \S\ref{sec:rules} cannot see.

\section{Further Base-Only Interventions}
\label{app:ablations}

\begin{figure}[h]
    \centering
    \includegraphics[width=1.0\linewidth]{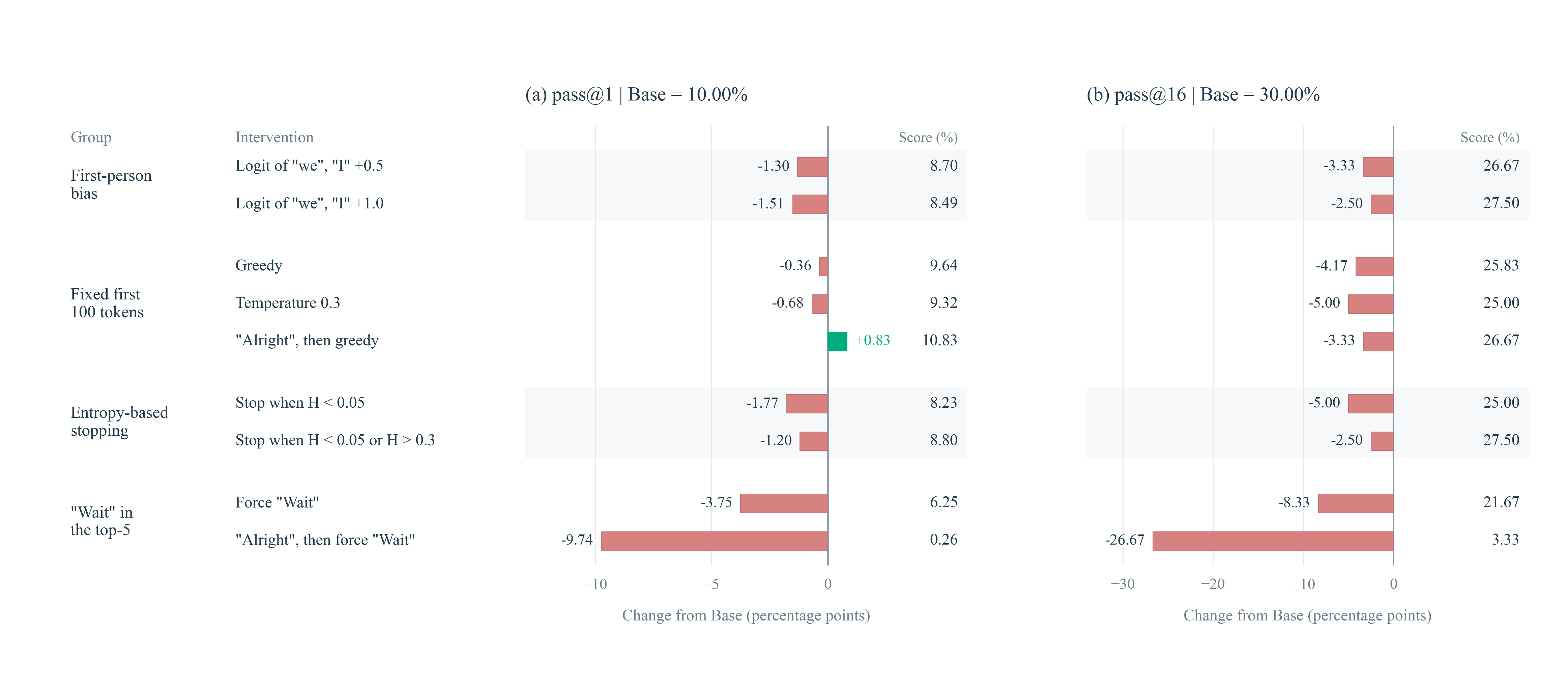}
    \caption{Further interventions on the base model, evaluated on 120 AIME problems with 16 samples per problem. Bars show changes in (a) pass@1 and (b) pass@16 relative to the base model (10.00\% and 30.00\%, respectively), in percentage points. Red indicates a decrease and green an increase. The score columns give the absolute pass@$k$ values (\%). The two panels use different horizontal-axis ranges.}
    \label{fig:intervention}
\end{figure}

Section~\ref{sec:rules} summarizes several interventions besides the ``Wait'' and ``Re'' rules. Figure~\ref{fig:intervention} compares their effects on pass@1 and pass@16. All interventions act on the base model alone and are evaluated on the 120 AIME problems with 16 samples per problem. The bars show changes relative to the base model, whose pass@1 and pass@16 are 10.00\% and 30.00\%, respectively; the score columns give the absolute values.

These interventions do not reproduce the improvements achieved by \rlrep{}. Most lower pass@1, and all lower pass@16. The first-person bias lowers pass@1 by 1.3--1.5 percentage points, within the variation between independent base runs (\S\ref{sec:exp-years}). Greedy and low-temperature decoding of the first 100 tokens leave pass@1 close to the base level while lowering pass@16. In the greedy condition, all 16 samples share the same initial 100 tokens. Adding ``Alright'' before the greedy prefix raises pass@1 by about 1.2 points over the greedy prefix alone, in line with the opening effect of \S\ref{sec:firsttoken}. This setting reaches 10.83\% pass@1, but its pass@16 remains below the base model at 26.67\%. The entropy rule with a single low threshold fires in only 13 of the 1{,}920 responses, and the two-sided rule in 301.

\paragraph{How often ``Wait'' is inserted.}
The top-5 rule inserts ``Wait'' 1.64 times per response on average, and 76.7\% of the responses receive no insertion at all. Once the rule fires it tends to fire again, often four to ten times in the same response. Accuracy falls with the number of insertions: 7.95\% (117 of 1{,}472) for responses with none, 1.81\% (3 of 166) for one to five, and 0 for the 282 responses with six or more. On the 36 problems the base model can solve, responses without an insertion are correct 26.8\% of the time and responses with one only 2.0\%. The base model writes ``Wait'' 0.30 times per response on its own, while \rlrep{} writes it 5.74 times and the RL model about 5.7 times and benefits from it, so the number of ``Wait''s is not what separates helpful from harmful use.

\section{Additional Details on Accuracy and Coverage}
\label{app:passk}

Section~\ref{sec:passk} presents the large-$k$ results,
with the Qwen3-4B curves in Figure~\ref{fig:passk}
and results for both Qwen model sizes in
Table~\ref{tab:passk-main}.
This appendix describes the response-level mixture
and the temperature sensitivity of the Qwen3-4B
RL baseline.

\paragraph{Response-level mixture.}
For each problem, every attempt independently selects
the RL model with probability 0.92 and the base model
with probability 0.08.
The selected model alone generates the complete response,
using the evaluation prompt of
Appendix~\ref{app:prompts}, temperature 0.6,
top-$p$ 0.95, and a maximum of 8{,}192 generated tokens.
No generated prefix is shared between attempts.

The mixture therefore combines the models across
complete responses, whereas \rlrep{} combines their
token choices within each response.
The weight 0.92 is chosen to approximately match
\rlrep{} at pass@1: the independently generated
mixture obtains 19.00\%, compared with 18.98\%
for \rlrep{}.
Its pass@16 and pass@256 are 38.78\% and 61.45\%,
respectively.
Figure~\ref{fig:passk}a connects these three reported
budgets with a dashed line.
The paired problem-level bootstrap interval for the
pass@256 difference is given in Table~\ref{tab:cis}.

\subsection{Temperature sensitivity of the RL baseline}
\label{app:temperature}

\begin{figure}[!htbp]
\centering
\includegraphics[width=\linewidth]
{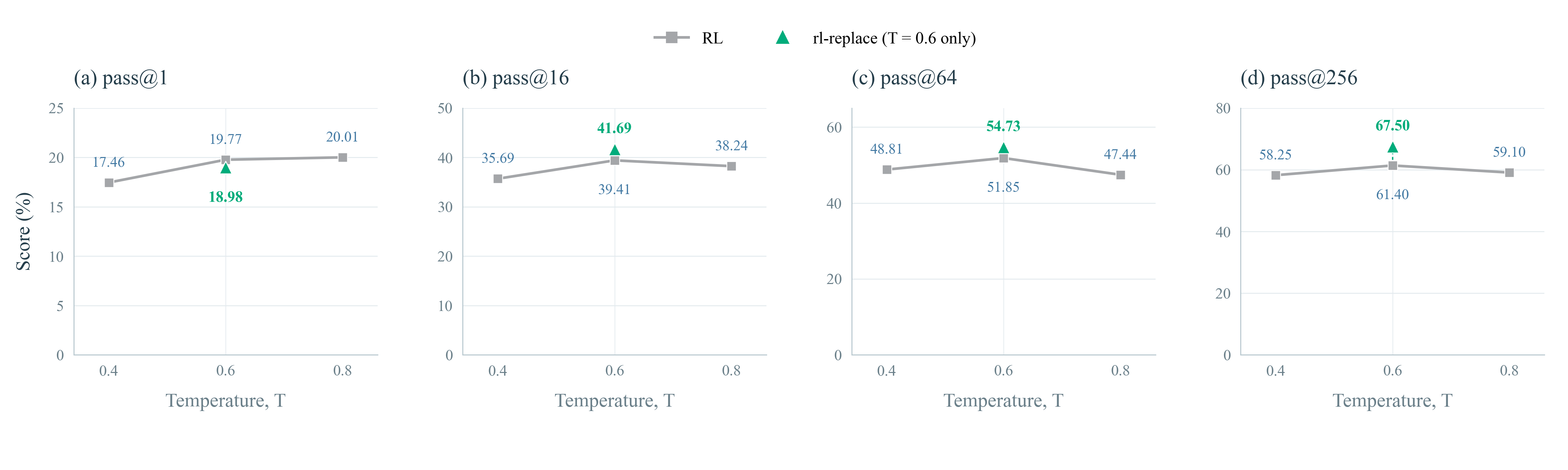}
\caption{Temperature sensitivity of the Qwen3-4B
RL baseline on all 120 AIME problems:
(a) pass@1, (b) pass@16, (c) pass@64, and
(d) pass@256 (\%), with top-$p$ fixed at 0.95.
Grey lines with square markers show the RL model
at temperatures 0.4, 0.6, and 0.8;
green triangles show \rlrep{} at temperature 0.6.
For \rlrep{}, temperature applies to base-model
sampling at agreement positions, while disagreements
use the RL model's greedy token.
Each panel starts at zero and uses its own
vertical-axis range.}
\label{fig:temp}
\end{figure}

We evaluate the Qwen3-4B RL model at temperatures
0.4, 0.6, and 0.8 on all 120 AIME problems.
Top-$p$ is fixed at 0.95, and all conditions use the
same evaluation prompt and 8{,}192-token output limit.
Figure~\ref{fig:temp} reports pass@1, pass@16,
pass@64, and pass@256.

The original temperature of 0.6 yields the highest
RL pass@256 among the tested settings, at 61.40\%.
Lowering the temperature to 0.4 gives 58.25\%,
while increasing it to 0.8 gives 59.10\%.
\rlrep{}, evaluated at temperature 0.6, reaches
67.50\%, exceeding the highest observed RL value
by 6.10 percentage points.
For \rlrep{}, temperature applies only to base-model
sampling at agreement positions; disagreement
positions use the RL model's greedy token.

This comparison is limited to the three tested
RL temperatures at a fixed top-$p$.
It does not characterize other decoding settings
or the temperature sensitivity of \rlrep{}.

\section{Qualitative Analyses}
\label{app:cases}

\subsection{The opening on the hardest problems}
\label{app:cases-opening}

Section~\ref{sec:firsttoken} measures the first token on all 120 problems. Here we look at the hardest ones: the eight problems on which both the base model and the RL model fail all 16 samples while \rlrep{} solves at least one. Table~\ref{tab:hard-opening} counts correct responses out of $8\times16=128$.

\begin{table}[h]
\centering
\caption{Correct responses on the eight problems that neither the base model nor the RL model solves in 16 samples (128 responses per condition).}
\label{tab:hard-opening}
\small
\begin{tabular}{@{}lcc@{}}
\toprule
Condition & Correct responses & Problems solved \\
\midrule
Base & 0 / 128 & 0 / 8 \\
Base + ``Alright'' & 2 / 128 & 2 / 8 \\
Base + ``Alright, I'' & 3 / 128 & 3 / 8 \\
RL model & 0 / 128 & 0 / 8 \\
\rlrep{} & 12 / 128 & 8 / 8 \\
\bottomrule
\end{tabular}
\end{table}

Forcing the opening solves two or three of these problems once each, while \rlrep{} solves all eight. Because the problems were selected by \rlrep{}'s success, its count is biased upward: with 256 samples its accuracy on them falls from 9.4\% to 3.6\%, so we neither compute a recovery rate nor count these problems as evidence that \rlrep{} solves problems beyond the RL model's reach. The comparison only shows that on the hardest problems the opening alone does very little, in line with \S\ref{sec:firsttoken}.

\subsection{Two worked examples}
\label{app:cases-example}

We show parts of two \rlrep{} responses. Tokens in bold are chosen by the RL model, at positions where the greedy choices of the two models differ; every other token is sampled from the base model.

\paragraph{A pause before an overlap check.}
AIME 2025 Problem~4 asks for the number of ordered pairs $(x,y)$ of integers between $-100$ and $100$, inclusive, with $12x^2-xy-6y^2=0$ (answer 117). The base model fails all 16 samples and never answers 117; its most common answer is 168 (6 samples). \rlrep{} answers correctly in 10 samples and the RL model in 15. The base model can factor the equation as $(3x+2y)(4x-3y)=0$ and so find the two families of solutions, but it never checks whether they overlap, and two of its samples answer 118.

The \rlrep{} response below has 2{,}085 tokens, of which the RL model chooses 84 (4.0\%). It counts 51 pairs in one family and 67 in the other. At the next position the base model's greedy token is ``\#\#\#'', which opens a new section header, and the RL model's token is ``Wait'', the only ``Wait'' in the response:

\begin{tcolorbox}[colback=gray!4,colframe=gray!50,boxrule=0.4pt,arc=2pt,left=5pt,right=5pt,top=3pt,bottom=3pt,fontupper=\small]
From Case 1, we have 51 \textbf{ordered} pairs. From Case 2, we have 67 ordered pairs.\\
Total number of ordered pairs: $51 + 67 = 118$

\textbf{Wait}, but I need to check if there's any overlap between the two cases\textbf{.} That is, are there any pairs $(x, y)$ that satisfy both \textbf{cases}?

\textbf{\#\#\#} Checking for Overlapping Solutions

\textbf{Let}'s see if there\textbf{'s} any $(x, y)$ that can be expressed \textbf{both} as $(3k, 4k)$ and as $(-2m, 3m)$. [\ldots]

So, the only overlapping solution is $(x, y) = (0, 0)$.

\textbf{Therefore}, the total number of \textbf{unique} ordered pairs is: $51 + 67 - 1 = 117$
\end{tcolorbox}

The rest of the check is mostly the base model's own text: it solves the two equations, finds that only $(0,0)$ lies in both families, and answers 117. This is the \#\#\#$\to$Wait pattern of \S\ref{sec:rules}.

\paragraph{A doubt that is not followed through.}
The second problem shows the same kind of replacement on a problem that no condition solves. Points $A,B,C,D$ lie on the hyperbola $\frac{x^2}{20}-\frac{y^2}{24}=1$ and form a rhombus whose diagonals meet at the origin; the task is to find the greatest real number that is less than $BD^2$ for all such rhombi (AIME 2024 Problem~9, answer 480). The base model, \rlrep{} and the RL model each fail all 16 samples, and 80 is the most common answer in each. Base responses open with ``We are given the hyperbola\ldots'' or ``To solve the problem, we'll follow these steps:'' and list steps at once; one of them concludes that $BD^2$ is unbounded.

In the \rlrep{} response below (1{,}611 tokens, 76 chosen by the RL model, 4.7\%), the first token is ``Alright''. The response lists the properties of a rhombus, including that its diagonals bisect each other at right angles, and then places the diagonals on the coordinate axes. At a position where the base model's greedy token is ``So'', the RL model's token is ``Wait'':

\begin{tcolorbox}[colback=gray!4,colframe=gray!50,boxrule=0.4pt,arc=2pt,left=5pt,right=5pt,top=3pt,bottom=3pt,fontupper=\small]
\textbf{Alright}, I have this problem about a hyperbola and a rhombus\textbf{.} Let's try to understand \textbf{it first}. [\ldots]

Since the diagonals \textbf{of} the rhombus intersect at the origin \textbf{and bis}ect each other, the \textbf{rh}ombus is \textbf{centered} at the origin. Let's denote the diagonals as $AC$ and $BD$, \textbf{with} $\boldsymbol{AC}$ along the $x$-axis and $BD$ along the $y$-axis for simplicity.

\textbf{Wait}, \textbf{is} that \textbf{necessarily the} case? The \textbf{problem says} the diagonals intersect at the origin, but it doesn't specify their \textbf{directions}. \textbf{Maybe I need} to \textbf{think} more carefully [\ldots]
\end{tcolorbox}

The doubt is correct, since the diagonals need not lie on the axes, but the response keeps the assumption, obtains $BD^2=960$, and answers 959. In the first example the base model carries the check through; here it does not.

\subsection{Where the RL model disagrees most}
\label{app:cases-disagree}

To see what the strongest disagreements look like, we inspected base-model responses on a set of 1{,}324 MATH problems at the positions where the RL model assigns the sampled token a much lower log-probability than the base model does. Two kinds of context recur. The first is a template opening such as ``Let's solve the problem step by step'' or ``First, let's consider\ldots'', where the base model itself is often unsure of its token and the RL model puts most of its probability on other tokens, in line with the opening change of \S\ref{sec:firsttoken}. The second is a premature conclusion in the middle of a derivation, such as ``So, the roots are:'' before the roots have been derived; connectives such as ``Given that\ldots'' or ``Since $a<b$\ldots'' are marked in the same way, less strongly. At an equals sign or a \LaTeX{} command the RL model usually agrees with the token itself, and the disagreement concerns what follows: the value on the right of the equals sign, or how an expression is written, such as the grouping when completing the square. These observations are qualitative and are not used for any number in the main text.

\end{document}